\documentclass[10pt,twocolumn,letterpaper]{article}
\usepackage[letterpaper,left=0.8in,right=0.8in,top=0.85in,bottom=0.85in,columnsep=0.25in]{geometry}
\usepackage[T1]{fontenc}
\usepackage{times}
\usepackage{microtype}
\usepackage{xspace}
\usepackage{graphicx}
\usepackage{amsmath,amssymb}
\usepackage{booktabs}
\usepackage{multirow}
\usepackage{array}
\usepackage[dvipsnames]{xcolor}
\usepackage[numbers,sort&compress]{natbib}
\usepackage[format=plain,labelfont=bf,labelsep=colon,font=small]{caption}
\usepackage[font=footnotesize,skip=3pt]{subcaption}
\usepackage[shortlabels,inline]{enumitem}
\usepackage{etoolbox}
\usepackage{afterpage}
\usepackage[hyphens]{url}
\definecolor{linkblue}{rgb}{0.08,0.33,0.62}
\usepackage[breaklinks,colorlinks,allcolors=linkblue]{hyperref}
\usepackage[capitalize,noabbrev]{cleveref}
\crefname{figure}{Figure}{Figures}\Crefname{figure}{Figure}{Figures}
\crefname{table}{Table}{Tables}\Crefname{table}{Table}{Tables}
\crefname{section}{Section}{Sections}\Crefname{section}{Section}{Sections}
\crefname{subsection}{Section}{Sections}\Crefname{subsection}{Section}{Sections}
\crefname{subsubsection}{Section}{Sections}\Crefname{subsubsection}{Section}{Sections}
\crefname{equation}{Equation}{Equations}\Crefname{equation}{Equation}{Equations}
\graphicspath{{fig/}}
\setlist{itemsep=1pt,topsep=3pt}
\makeatletter
\renewcommand\section{\@startsection{section}{1}{\z@}{-2.0ex \@plus -.5ex \@minus -.2ex}{1.0ex \@plus .2ex}{\normalfont\large\bfseries}}
\renewcommand\subsection{\@startsection{subsection}{2}{\z@}{-1.6ex \@plus -.4ex \@minus -.2ex}{0.7ex \@plus .2ex}{\normalfont\normalsize\bfseries}}
\renewcommand\paragraph{\@startsection{paragraph}{4}{\z@}{1.4ex \@plus .5ex \@minus .2ex}{-1em}{\normalfont\normalsize\bfseries}}
\renewcommand\@maketitle{\newpage\null\vskip 0.6em
  \begin{center}{\LARGE\bfseries \@title \par}\vskip 1.1em
  {\large \lineskip .5em \begin{tabular}[t]{c}\@author\end{tabular}\par}\end{center}\par\vskip 1.2em}
\makeatother
\renewenvironment{abstract}{\par\begin{center}\bfseries Abstract\end{center}\vspace{-0.8ex}\noindent\ignorespaces}{\par\vspace{1.2ex}}

\newcommand{\rhythm}{VAE Rhythm\xspace}
\newcommand{\engine}{VAE-pair engine\xspace}
\newcommand{\method}{CaDet\xspace}
\newcommand{\methodgf}{CaDet-GF\xspace}
\newcommand{\plugin}{VAE Rhythm plug-in\xspace}
\newcommand{\pairs}{VAE pairs\xspace}
\newcommand{\Cfour}{C_4}
\newcommand{\Ceight}{C_8}
\newcommand{\Ppk}{P_{\mathrm{pk}}}
\newcommand{\best}[1]{\textbf{#1}}

\newcommand{\ms}[2]{#1\textsubscript{\tiny$\pm$#2}}
\newcommand{\dsc}[1]{\textcolor{gray}{\scriptsize #1}}
\newif\ifrobust

\robusttrue
\title{VAE Rhythm:\\Temporal Decoding Traces for AI-Generated Video Detection}
\author{Ke Sun\qquad Shuwei Li\qquad Haiyang Mei\qquad Mike Zheng Shou\textsuperscript{\dag}\\[0.7ex]{\normalsize Show Lab, National University of Singapore}\\[0.6ex]{\small Project page: \href{https://showlab.github.io/vae-rhythm}{showlab.github.io/vae-rhythm}}}
\date{}
\begin{document}
\maketitle
{\renewcommand{\thefootnote}{}\footnotetext{\textsuperscript{\dag}Corresponding author.}}
\suppressfloats[t]
\begin{abstract}
Visible artifacts are an unreliable cue for detecting generated videos, because newer generators learn to remove them.
We instead detect a trace left by the generation pipeline itself.
Most video generators decode latents with a 3D causal VAE that groups consecutive frames in time, so the fidelity of generated frames rises and falls periodically with their position in a group.
We call this periodic trace \textbf{\rhythm}.
Real videos are captured frame by frame and do not have it.
To reveal it without training, we propose a \emph{phase scan}.
The scan reconstructs a clip with an off-the-shelf VAE from successive start offsets and measures the periodicity of the reconstruction error across these offsets.
We find \rhythm in most of the 39 variants of 31 generators we test, including the closed-source Sora, Kling~3.0 and Seedance~2.0.
We use the rhythm as a zero-shot score, as a plug-in for existing detectors and as a data engine that pairs real videos with their VAE reconstructions to train our detector \method.
On DF26, the training-free rhythm score outperforms all 15 retrained baselines, and as a plug-in it improves all 19 tested detectors by 6.7\% to 23.7\% AUROC.
\method achieves the best AUROC on AIGVDBench, DF26 and three 2026 closed-source models, and its variant trained without any generated video still exceeds 90\% on AIGVDBench.
\looseness=-1
\end{abstract}

\begin{figure}[!t]
\centering
\includegraphics[width=0.984\columnwidth]{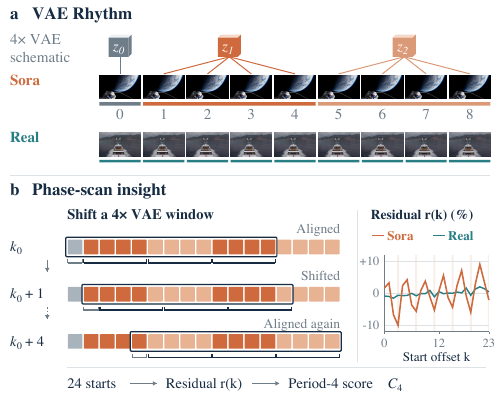}
\caption{\textbf{VAE Rhythm.} (a)~A $4\times$ causal VAE decodes frames in groups, illustrated on Sora frames. (b)~As the VAE window shifts, its groups realign with those of the clip every four starts. The residual $r(k)$ of a typical Sora clip therefore oscillates, while that of a typical real clip stays nearly flat.}
\label{fig:teaser}
\end{figure}

\section{Introduction}
\label{sec:intro}

Video generators such as Sora~\cite{sora,sora2}, Veo~\cite{veo2,veo31}, Kling~\cite{kling} and Seedance~\cite{seedance1,seedance2report} now produce clips that people can no longer reliably tell from camera recordings.
On DF26, human viewers identify generated clips with near-chance accuracy~\cite{df26}.
Preference tuning~\cite{ipoc} lets each new model remove the flicker and deformation that viewers notice, so detectors built on such visible artifacts chase a moving target.
A durable cue should lie in the \textbf{\textit{perceptual blind spot}} of video generation.
Traces in this blind spot are invisible to humans, so developers have no reason to remove them.

To find such traces, we look at how generated videos are formed.
We observe that most open-source~\cite{wan,hunyuan,ltx,cosmos} and closed-source~\cite{seedance1,minimaxh3} video generators decode latents into frames with a 3D causal VAE~\cite{ldm,dit,cogvideox}.
Specifically, to reduce computation, the VAE compresses time by grouping consecutive frames into shared latents.
In the standard $4\times$ scheme of \cref{fig:teaser}a, the first frame is encoded alone and each later group of four frames shares one latent.

This grouping is invisible to viewers, but it fixes the position of each generated frame within a group.
\emph{Can we detect this grouping from the pixels of a video without access to its generator?}
To investigate this question, we reconstruct generated videos with an off-the-shelf video VAE and shift the frame at which the reconstruction starts.
We find that the reconstruction error rises and falls periodically with this shift, and the period equals the group size.
Because causal 3D convolutions treat the positions within a group differently, the fidelity of a generated frame depends on its position in its group~\cite{ivvae}.
The reconstruction error is therefore lowest whenever the groups of the two VAEs align.
We call this periodic phenomenon \textbf{\rhythm}.
A camera captures each frame independently, so real videos lack it.

We measure \rhythm with a training-free \emph{phase scan}, shown in \cref{fig:teaser}b.
A frozen $4\times$ VAE reconstructs a $W$-frame window of the clip from each start offset $k=0,\dots,K-1$.
For each reconstructed frame, we subtract the mean error of that frame over all windows that contain it, which removes the effect of frame content.
Averaging these differences within the window gives a residual $r(k)$ for offset $k$.
The rhythm score is the share of the variance of $r(k)$ at the period of the rhythm.
Earlier reconstruction-based detectors use the level of the error~\cite{dire,aeroblade,lare2,divid}, whereas the phase scan uses only its periodicity.

We run the phase scan on \textbf{39 variants of 31 generators} released from 2023 to 2026.
The rhythm appears in both open-source and closed-source models.
\Cref{fig:rhythm-curves} shows it in Sora, Kling~3.0, Grok Imagine~\cite{grok} and Seedance~2.0.
Compared with real clips, the rhythm is significantly stronger in 30 of 36 matched variants.
Different generated clips also oscillate in phase.
Their error minima tend to fall at the same start offsets, because every generated video starts its first group at its first frame.
Real clips show no such common offset.
In control experiments, a VAE round trip induces the rhythm in real clips, whereas H.264~\cite{h264} re-encoding does not.

\rhythm is thus invisible to viewers but measurable from pixels alone, so it lies in the perceptual blind spot of video generation.
We exploit it for detection in three ways.
First, the rhythm score itself is a \emph{zero-shot} detector that needs no training.
Second, the \emph{\plugin} adds the rhythm score to the score of any existing detector when the rhythm is stronger than in typical real videos, so the detector needs no retraining.
Third, the \emph{\engine} builds training data from real videos alone.
It passes each real clip through a frozen VAE and post-processes the clip and its reconstruction identically, so the two differ only by VAE decoding.
We build our own detector on these pairs and call it \method, which stands for Cadence Detector.
\method learns VAE decoding traces from the pairs in an artifact path and adds a semantic path that never sees them.
Its generator-free variant \methodgf trains only the artifact path on real videos and pairs, so it never sees generated videos.

All three uses improve detection.
Without training, the zero-shot score reaches 76.2\% AUROC on DF26 and beats the best of 15 retrained baselines by 5.8\%.
The plug-in improves all 19 tested detectors on DF26 by 6.7\% to 23.7\%.
\method is best on AIGVDBench, DF26 and three 2026 closed-source models, and \methodgf reaches 90.9\% and 91.9\% on AIGVDBench's open- and closed-source generators.

Our contributions are:
\begin{itemize}
    \item We identify \textbf{\rhythm} as a periodic trace of temporal grouping in video VAEs and propose a training-free phase scan to measure it.
    \item We analyze \rhythm in 39 variants of 31 generators and show that it is common in open- and closed-source models and originates in VAE decoding.
    \item We use \rhythm as a zero-shot score, a plug-in and a VAE-pair engine, and improve detection on AIGVDBench, DF26 and 2026 closed-source models.
\end{itemize}

\section{Related Work}
\label{sec:related}

\paragraph{Synthetic media and video detection.}
Following face and image forgery benchmarks~\cite{ffpp,genimage,diffusionface}, video benchmarks include GenVideo~\cite{demamba}, DVF~\cite{mmdet}, GenVidBench~\cite{genvidbench}, AIGVDBench~\cite{aigvdbench}, ViF-Bench~\cite{skyra}, GenBuster~\cite{busterx,busterxpp}, CoCoVideo~\cite{cocovideo} and DF26~\cite{df26}.
Image and face detectors study spatial, frequency and attention cues~\cite{cnnspot,f3net,gramnet,fredect,fusing,sia,npr}.
Others use domain-general, contrastive or continual learning~\cite{ltw,dcl,sbi,ucf,hdp}, or adapt frozen vision backbones~\cite{clip,dinov2} through intermediate features, subspaces or transferable concepts~\cite{univfd,fatformer,rine,effort,forgelens,forensicconcept}.
Video detectors build on action-recognition backbones~\cite{timesformer,uniformerv2,videomae,xclip}, spatiotemporal models~\cite{tall,demamba,decof,aigvdet,unite,waverep,nativescale}, training-free feature statistics~\cite{d3,restrav,stall,split,nsgvd} and multimodal LLMs that reason about images and videos~\cite{vlffd,aigiholmes,busterx,busterxpp,vidguard,skyra,videoveritas}.
Audits also document format and metadata confounds in existing benchmarks~\cite{fakejpeg,waverep,vidguard,motionbias,vidaudit}.
We instead measure a temporal trace of VAE decoding that many generator families share.

\paragraph{Generator-derived cues and video VAEs.}
Reconstruction error is an established cue for generated images~\cite{dire,aeroblade,lare2,drct,fire}, and diffusion models can also guide detector features~\cite{diffusionfake}.
Video detectors use reconstruction error~\cite{divid,reconfuse} and inversion noise~\cite{dbinds}, and concurrent work probes flow trajectories~\cite{trace}.
Image~\cite{aligned,bfree,drct} and video detectors~\cite{waverep,g2vd} also train on pairs of real data and their autoencoder reconstructions, with the reconstructions labelled as generated.
Video VAEs compress time with causal 3D convolutions~\cite{magvit2,cogvideox,wan,hunyuan,ltx,cosmos,pyramid,easyanimate}.
IV-VAE~\cite{ivvae} reports unequal quality within frame groups, and WF-VAE~\cite{wfvae} reduces seams between chunks.
We instead measure the \emph{periodicity} of reconstruction error across temporal alignments.
We also build VAE pairs from real videos with shared post-processing, so the VAE is the only difference within a pair.

\section{Understanding VAE Rhythm}
\label{sec:analysis}
\begin{figure*}[t]
\centering
\begin{minipage}[t]{0.48\textwidth}
\centering
\includegraphics[width=0.96\linewidth]{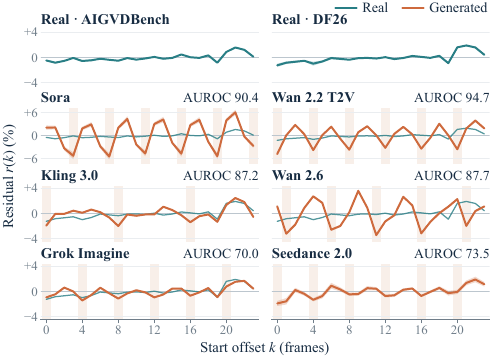}
\caption{\textbf{VAE Rhythm in real and generated videos.} Mean residuals $r(k)$ over 50--300 clips. Pale bands mark the period of each panel, and thin lines show the real clips of the same benchmark. Panel AUROCs are the zero-shot scores of \cref{sec:zeroshot}. \cref{fig:rhythm-curves-full} in the supplement shows all 39 variants.}
\label{fig:rhythm-curves}
\end{minipage}\hfill
\begin{minipage}[t]{0.48\textwidth}
\centering
\includegraphics[width=0.96\linewidth]{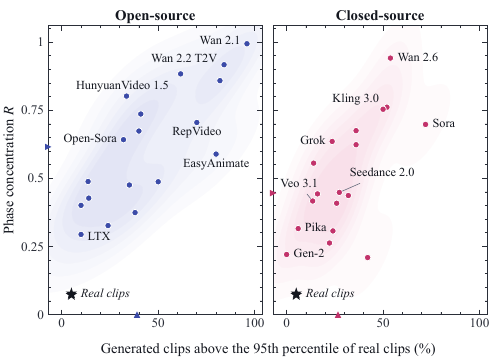}
\caption{\textbf{Rhythm strength and phase, real vs.\ generated.} Each point is one window-matched variant. It plots the phase concentration $R$ against the share of clips above the 95th percentile of real clips. Shading shows the density of variants, and triangles mark the medians.}
\label{fig:rhythm-stats}
\end{minipage}
\end{figure*}

\paragraph{Mechanism.}
A causal video VAE with temporal compression factor $T$ maps $1+Tn$ frames to $1+n$ latent steps~\cite{magvit2,cogvideox,wan}.
It encodes the first frame separately, and each later latent step expands into a group of $T$ output frames.
The group boundaries at frames $1,1+T,1+2T,\dots$ form a temporal lattice.
Causal convolutions can span several groups, but they treat the positions in this lattice differently, so reconstruction quality is unequal within a group~\cite{ivvae}.
We call the resulting periodic signature \emph{\rhythm}.
Camera acquisition has no such lattice, although codecs and frame-rate conversion can add their own periodic structure, which \cref{sec:controls} in the supplement examines.

\subsection{Revealing the rhythm by phase scan}
\label{sec:scan}
The phase scan in \cref{fig:teaser}b uses the frozen $4\times$ causal VAE of Wan2.1~\cite{wan} by default.
\Cref{fig:scan} in the supplement shows the full error table and the baseline subtraction.
For each start offset $k=0,\dots,K-1$, we reconstruct a $W$-frame window and record the RGB mean-squared error $e(k,j)$ of its $j$th frame, with pixels normalized to $[-1,1]$.
By default $K=24$ and $W=13$, and 33-frame clips use $W=9$.
When the scan and source lattices align, reconstruction is easier, so the error is periodic across offsets.

The window content also changes with $k$.
To remove this variation, let $\bar e(t)$ be the mean error of absolute frame $t$ over all windows that cover it.
We exclude window position $j=0$ from this mean and from the residual, because the VAE treats its first frame separately.
The dimensionless profile is
\begin{equation}
r(k)=\frac{1}{(W-1)\bar E}\sum_{j=1}^{W-1}\big[e(k,j)-\bar e(k+j)\big],
\label{eq:profile}
\end{equation}
where $\bar E$ is the mean of the full error table.
This removes the content baseline of each frame but not every effect of post-processing.
$\Cfour$ and $\Ceight$ measure the fraction of residual variance in the period-4 and period-8 harmonic combs.
The period-agnostic score $\Ppk$ is the largest normalized periodogram bin for periods from 2.5 to 9, after $4\times$ zero-padding.
This range includes period 5, which appears when a four-frame pattern at 24\,fps is delivered at 30\,fps.
It excludes period-2 resampling artifacts and long-period drift.
The phase $\varphi$ is the argument of the period-4 or period-8 Fourier coefficient.
Its concentration across $N$ clips is $R=|N^{-1}\sum_i\exp(\mathrm{i}\varphi_i)|$, which is 0 for dispersed phases and 1 for perfect alignment.

\subsection{VAE Rhythm across generators}
\label{sec:evidence}
\begin{figure*}[t]
\centering
\includegraphics[width=\textwidth]{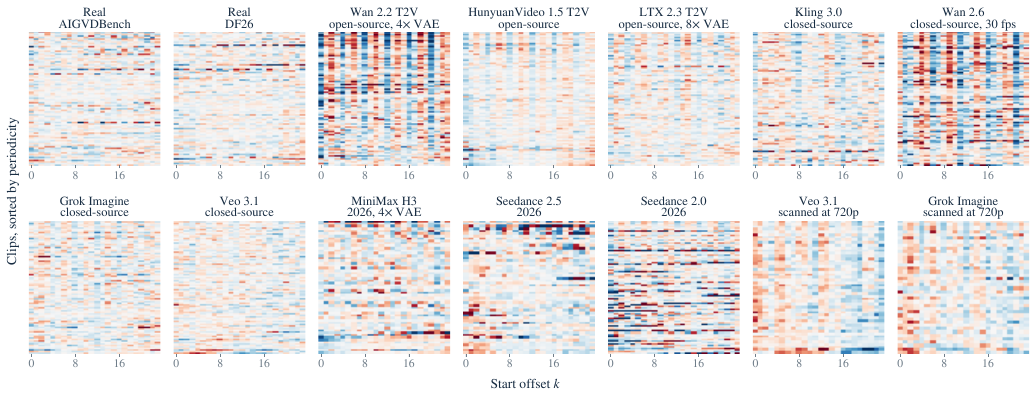}
\caption{\textbf{Barcode view.} Residual profiles of up to 100 clips per group with one clip per row. Red is high and blue is low.}
\label{fig:supp-barcode}
\end{figure*}
Our census covers all 39 generator variants that support 24 offsets: 26 from AIGVDBench, 10 from DF26~\cite{wan22,hunyuan15,ltx2,wan26}, MiniMax~H3, and Seedance~2.0 and 2.5.
The five 16-frame 2D-VAE variants in AIGVDBench~\cite{animatediff,svd,seine,videocrafter2} are too short for this scan.

\paragraph{Qualitative evidence.}
\Cref{fig:rhythm-curves} shows the mean profiles $r(k)$ of the two real cohorts and of six generators that we selected for signal strength and prominence.
\Cref{fig:rhythm-curves-full} in the supplement shows all 39 variants, including weak and negative cases.
We average the profiles as measured, without phase alignment, smoothing or subtraction of the real mean.
Repeated peaks and troughs survive this averaging over unrelated clips in open-source and closed-source models alike.
The period is four frames in most families, eight in LTX and Kling~3.0, and about five in Wan~2.6.
The real means show no repeating pattern, only a late-offset excursion that the generated means share.

\paragraph{Quantitative evidence.}
We compare the 36 variants with window-matched scans against the real clips of their own benchmark.
In \cref{fig:rhythm-stats}, real clips sit in the lower-left corner, and most open-source and closed-source variants lie far above and to the right.
\Cref{tab:rhythm-stats} in the supplement lists every value.
The rhythm statistic of the median variant is 2.2 times that of real clips.
A threshold that only 5\% of real clips exceed is exceeded by 34\% of its clips.
A one-sided Mann--Whitney test with Holm correction finds a significant difference at $p<0.05$ for 30 of the 36 variants.
The phases also separate the two classes.
\Cref{fig:supp-barcode} stacks the residual profiles of single clips, and generated clips form aligned bands that real clips lack.
After we remove the mean real profile, the phases of real clips are consistent with a uniform distribution, with $R$ near 0.08.
The median variant reaches $R=0.52$, as expected when the groups of every clip start at its first frame.
Gen-2 and Pika are exceptions, and their rhythm statistic is lower than that of real clips.
Jimeng, Wan (web) and two LTX variants do not reach significance with 50 clips each.

\paragraph{Further tests.}
\Cref{sec:supp-analysis} in the supplement collects our tests of the origin and limits of the rhythm.
A VAE round trip induces the rhythm in real clips, whereas matched x264 re-encoding does not.
After a round trip through the Wan2.1 VAE, the period-4 score separates real clips from their reconstructions with 90\% AUROC.
Real clips re-encoded with the x264 settings of Grok stay at about 49\%.
Removing two leading frames shifts its phase by two positions.
The rhythm weakens in nearly static clips and when the scan resolution is far below the generation resolution.
For real clips reconstructed by a VAE, the median period-4 score falls from 0.64 to 0.33 when the scan resolution drops to about half the reconstruction resolution.
Scanning at native resolution instead of 256p raises the zero-shot AUROC of Sora from 72\% to 90\%.
We therefore scan each clip at its native resolution and downscale it only when its short side exceeds 480 pixels, a setting we call 480p.
For clips delivered with a 720-pixel short side, we also scan at that size and call this setting 720p.

\subsection{Zero-shot detection with the rhythm}
\label{sec:zeroshot}
\begin{figure*}[t]
\centering
\includegraphics[width=\textwidth]{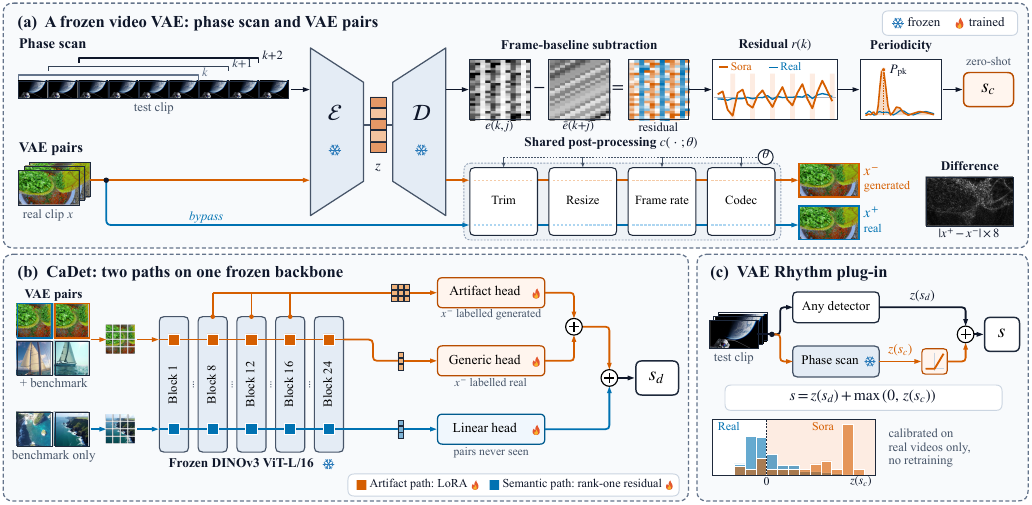}
\caption{\textbf{Overview.} (a)~A frozen video VAE has two uses. The periodicity of its phase scan gives the zero-shot score $s_c$, and its round trip creates the \pairs, which differ only by VAE decoding. (b)~On one frozen DINOv3 backbone, \method trains an artifact path with the pairs and a semantic path without them. (c)~The plug-in adds positive rhythm evidence to the score of any detector without retraining it. All plots, tables and the difference map are measured.}
\label{fig:method}
\end{figure*}

\begin{table}[t]
\centering
\caption{\textbf{Zero-shot AUROC of representative generators} (\%) using the rhythm statistic alone without training. \cref{tab:census} in the supplement lists all 39 variants.}
\label{tab:zeroshot}
\scriptsize
\setlength{\tabcolsep}{5.2pt}
\begin{tabular}{@{}lc@{\hskip 14pt}lcc@{}}
\toprule
Open-source & 480p & Closed-source & 480p & 720p \\
\midrule
Wan2.1 T2V & 99.5 & Sora & 90.4 & 84 \\
Wan~2.2 T2V & 94.7 & Wan~2.6 & 87.7 & 96.0 \\
EasyAnimate I2V & 94.2 & Kling~3.0 & 87.2 & 88.2 \\
RepVideo & 89.3 & Luma & 79.3 & 89 \\
Open-Sora & 77.0 & Seedance~2.0 & 73.5 & 78.5 \\
HunyuanVideo~1.5 T2V & 76.7 & Grok Imagine & 70.0 & 79.4 \\
LTX~2.3 T2V & 76.5 & MiniMax~H3 & 67.3 & 78.9 \\
Pyramid-Flow I2V & 71.3 & Veo~3.1 & 57.9 & 73.4 \\
LTX T2V & 52.0 & Gen-2 & 30.7 & 26 \\
\bottomrule
\end{tabular}
\end{table}

The tests above compare groups of clips.
To check whether the rhythm also separates individual clips, we use it directly as a detection score without training, thresholds or calibration.
We scan each clip once with the frozen Wan2.1 VAE and score it with the share of residual variance at the period of its family.
This score is $\Cfour$ by default, $\Ceight$ for the eight-frame Kling~3.0 and LTX~2.3 T2V, and $\Ppk$ for the period-5 Wan~2.6.
We measure its AUROC against the real clips of the same benchmark, where 50\% is chance.

\Cref{tab:zeroshot} shows representative results, and \cref{tab:census} in the supplement lists all variants.
The score exceeds chance on 34 of the 36 window-matched variants.
As the mechanism predicts, it is strongest on open-source generators with a $4\times$ VAE.
It also works on closed-source models with unknown VAEs, such as Sora and Kling~3.0.
It is weaker on the $8\times$ LTX family, because our $4\times$ measurement VAE does not match it.
Most DF26 and 2026 generators improve at 720p, which is consistent with the scale dependence.
Gen-2 and Pika stay below chance, which matches their flat curves.
With the single statistic $\Ppk$ for all generators, the score reaches 76.2\% on DF26's ten generators.
The best retrained baseline on DF26 is Effort, and the score exceeds it by 5.8\%.
Supervised detectors train on AIGVDBench's own Open-Sora clips, so on AIGVDBench the score trails most of them.

\section{Method}
\label{sec:method}

\Cref{sec:analysis} shows that VAE decoding causes the rhythm and that an off-the-shelf VAE measures it across generators.
We therefore build on a frozen VAE and real videos rather than on any particular generator.
\Cref{fig:method} summarizes the method.

\subsection{The \engine}
\label{sec:engine}
The controls of \cref{sec:evidence} show that a VAE round trip alone induces the rhythm in real clips.
Sampling generators is costly and often impossible for closed models, so we create training data by passing real clips through a frozen VAE.
Let $x$ be a real clip and $s_G$ a rescaling to a reconstruction short side $G$.
Let $\mathcal{D}\circ\mathcal{E}$ be one round trip through a frozen video VAE and $c(\cdot;\theta)$ a post-processing chain with random parameters $\theta$.
The engine emits
\begin{equation}
x^{+}=c\big(s_G(x);\theta\big),\qquad x^{-}=c\big(\mathcal{D}(\mathcal{E}(s_G(x)));\theta\big),
\label{eq:pairs}
\end{equation}
labelled real and generated.
The two members share content and every random choice of the chain, so the VAE is their only difference.
Many benchmarks confound generation with format and metadata~\cite{fakejpeg,vidaudit}.
Within a pair, format and metadata are identical, so they cannot tell the two members apart.
The chain mimics the processing of online videos.
It trims a few leading frames and rescales the clip to a common delivery size.
It may also interpolate, duplicate or drop frames before it re-encodes the clip with x264 or x265 at a random quality.
\Cref{sec:supp-chain} in the supplement gives the parameters.
Both members share the chain, so the detector learns VAE traces that survive this processing.
Processing alone also never signals generation.
Each pair draws its VAE from three $4\times$ VAEs and one $8\times$ VAE, so the detector learns traces that several VAEs share.
Following the scale law in \cref{sec:supp-scale} of the supplement, we sample $G$ over several scales and let the chain upscale some clips.

\subsection{\method: two paths on one frozen backbone}
\label{sec:detector}
\begin{table*}[t]
\centering
\caption{\textbf{Main results} (AUROC, \%). All detectors are trained on AIGVDBench's Open-Sora and real clips and tested without retraining. $\pm$ is the standard deviation over 3 seeds, and bold marks the best per column.}
\label{tab:main}
\footnotesize
\setlength{\tabcolsep}{5.5pt}
\begin{tabular}{@{}llccc|c|cccc@{}}
\toprule
 & & \multicolumn{3}{c|}{AIGVDBench} & 2026 closed & \multicolumn{4}{c@{}}{DF26} \\
Method & venue & 20 open & 19 unseen & 11 closed & 3 models & all & open & closed & 2AFC \\
\midrule
\multicolumn{10}{@{}l}{\emph{Image forensics, frame-level}} \\
CNNSpot~\cite{cnnspot} & CVPR'20 & \ms{82.1}{1.2} & \ms{81.1}{1.2} & \ms{71.4}{2.4} & 78.0 & 53.5 & 65.5 & 35.5 & 61.0 \\
Gram-Net~\cite{gramnet} & CVPR'20 & \ms{82.4}{0.4} & \ms{81.4}{0.5} & \ms{67.0}{1.4} & 76.9 & 52.9 & 63.3 & 37.3 & 59.4 \\
FreDect~\cite{fredect} & ICML'20 & \ms{60.9}{1.8} & \ms{58.9}{1.9} & \ms{52.9}{1.2} & 60.8 & 60.8 & 59.0 & 63.4 & 62.4 \\
Fusing~\cite{fusing} & ICIP'22 & 67.0 & 65.3 & 51.5 & 64.9 & 46.0 & 53.3 & 35.0 & 46.4 \\
UnivFD~\cite{univfd} & CVPR'23 & \ms{77.3}{0.0} & \ms{76.1}{0.1} & \ms{65.4}{0.1} & 60.3 & 54.0 & 59.4 & 46.1 & 62.5 \\
NPR~\cite{npr} & CVPR'24 & \ms{73.9}{0.8} & \ms{72.5}{0.8} & \ms{60.7}{0.8} & 69.5 & 44.1 & 53.1 & 30.7 & 47.2 \\
\multicolumn{10}{@{}l}{\emph{Video classifiers}} \\
TimeSformer~\cite{timesformer} & ICML'21 & 74.3 & 73.0 & 90.7 & 86.4 & 67.0 & 61.2 & 75.8 & 73.4 \\
UniFormerV2~\cite{uniformerv2} & ICCV'23 & 81.4 & 80.4 & 91.2 & 64.9 & 50.6 & 47.5 & 55.3 & 52.4 \\
\multicolumn{10}{@{}l}{\emph{Video detectors}} \\
DeMamba~\cite{demamba} & 2024 & 80.3 & 79.3 & 78.1 & 71.6 & 61.4 & 60.8 & 62.3 & 67.2 \\
DeCoF~\cite{decof} & ICME'25 & 80.6 & 79.6 & 78.7 & 70.9 & 59.8 & 60.7 & 58.5 & 68.0 \\
ReStraV~\cite{restrav} & NeurIPS'25 & 52.2 & 50.0 & 48.6 & 62.7 & 57.4 & 59.7 & 53.9 & 55.6 \\
ForgeLens, setting 3~\cite{forgelens} & ICCV'25 & \ms{75.5}{1.8} & \ms{74.2}{1.9} & \ms{82.4}{5.9} & 74.3 & 55.6 & 52.7 & 60.1 & 60.0 \\
ForgeLens, setting 1~\cite{forgelens} & ICCV'25 & \ms{74.1}{2.8} & \ms{72.7}{3.0} & \ms{65.0}{1.9} & 63.6 & 46.6 & 56.9 & 31.3 & 50.2 \\
\quad setting-1 schedule, full data &  & 85.7 & 84.9 & 87.4 & 83.2 & 60.3 & 62.7 & 56.5 & 66.9 \\
Effort~\cite{effort} & ICML'25 & \ms{84.3}{1.2} & \ms{83.4}{1.3} & \ms{92.5}{1.9} & 81.6 & 70.4 & 68.4 & 73.4 & 82.6 \\
Native-Scale~\cite{nativescale} & ICLR'26 & \ms{73.4}{2.1} & \ms{72.0}{2.2} & \ms{68.0}{4.0} & 78.9 & 52.2 & 53.9 & 49.6 & 57.9 \\
\midrule
\multicolumn{10}{@{}l}{\emph{Ours}} \\
DINOv3-L LoRA, no pairs (same backbone) & & 68.9 & 67.3 & 81.2 & 79.3 & 62.1 & 60.0 & 65.2 & 70.4 \\
\methodgf: real videos + pairs & & \ms{90.9}{3.0} & \ms{90.5}{3.2} & \ms{91.9}{3.0} & 84.8 & 65.8 & 75.5 & 51.3 & 70.5 \\
\method: benchmark data + pairs, two paths & & \best{\ms{93.9}{0.6}} & \best{\ms{93.6}{0.6}} & \best{\ms{95.8}{0.7}} & \best{88.3} & \best{78.5} & \best{78.2} & \best{79.0} & \best{87.8} \\
\bottomrule
\end{tabular}
\end{table*}

Decoding traces may weaken or change when a closed-source generator uses an unknown VAE and post-processing, so cues that do not depend on the VAE are also needed.
\Cref{sec:exp-ablation} shows that one model does not easily learn both, because adding the \pairs to Effort~\cite{effort} hurts it on closed-source generators.
\method therefore attaches two paths to one frozen DINOv3 ViT-L/16~\cite{dinov3}.
The \emph{artifact path} learns decoding traces from the pairs, and the \emph{semantic path} never sees them.
Freezing the backbone keeps its pretrained features and is meant to limit overfitting.
Both paths share the frozen weights, so \method stores one backbone and two small trainable parts.

\paragraph{Artifact path.}
LoRA~\cite{lora} adapters on the attention projections feed two heads.
The \emph{generic head} reads the final class token and is trained with counterfactual labels.
Both members of every pair count as real, and only generator-produced clips count as generated.
\Cref{sec:supp-train} in the supplement shows that VAE reconstructions are smoother and less noisy than their sources.
A head that calls them generated may learn that smoothness means generation and then miss sharp closed-source videos.
The counterfactual labels make this appearance useless to the generic head.
The \emph{artifact head} reads class-token and patch-token statistics of intermediate layers because we expect these layers to keep more low-level detail.
It keeps the original labels to learn the decoding traces, and its gradient also updates the shared adapters.
The path score sums the two head scores after standardizing each on real videos.
\Cref{tab:ablation} in the supplement supports the two heads and the shared gradient.

\paragraph{Semantic path.}
Following Effort~\cite{effort}, this path freezes the principal singular subspace of every attention matrix and trains only the smallest singular component and a linear head.
Effort argues that the frozen subspace preserves pretrained knowledge and that the small trainable part limits overfitting.
The path trains frame by frame, and a clip's score is the logit of its mean frame probability.

\paragraph{Fusion and the generator-free variant.}
At test time, we standardize the score of each path by its mean and standard deviation on held-out real clips and sum the two to obtain $s_d$.
Both terms then count standard deviations above typical real videos, so we add them without fitting any weight on generated videos.
\methodgf trains only on real videos and their pairs.
The generic head and the semantic path would then lack generated examples.
\methodgf is therefore a single LoRA path with one linear head and the original labels, and it selects checkpoints on held-out pairs.

\subsection{The \plugin}
\label{sec:plugin}
\begin{figure*}[t]
\centering
\includegraphics[width=0.95\textwidth]{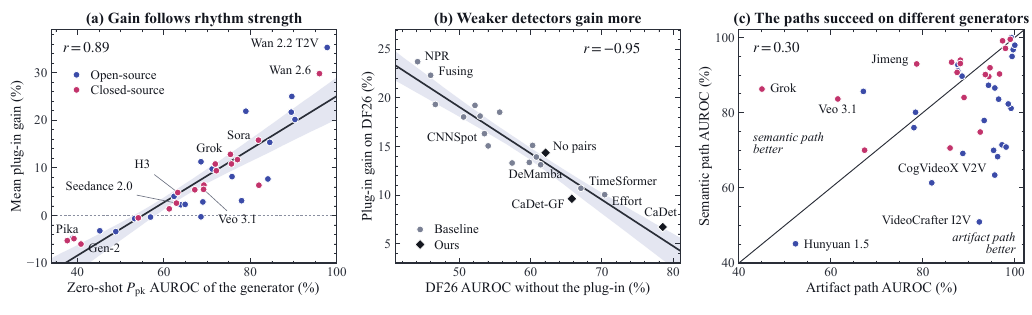}
\caption{\textbf{Analysis of the plug-in and the two paths.} (a)~Each dot is a generator. The plug-in gain averaged over the 19 detectors grows with the zero-shot AUROC of the generator. (b)~Each dot is a detector, and weaker detectors gain more on DF26. (c)~Each dot is an unseen generator, and the semantic path is better above the diagonal. Colors follow (a), and bands are bootstrap 95\% intervals of the fit.}
\label{fig:analysis}
\end{figure*}

\afterpage{\afterpage{\begin{table}[t]
\centering
\caption{\textbf{The \plugin on representative detectors} (AUROC, \%). Gray marks the change from the same detector without it, and bold marks the best per column. \cref{tab:plugin-full} in the supplement lists all 19.}
\label{tab:plugin}
\scriptsize
\setlength{\tabcolsep}{2.4pt}
\begin{tabular}{@{}lccccc@{}}
\toprule
 & \multicolumn{2}{c}{AIGVD} & 2026 & \multicolumn{2}{c}{DF26} \\
+ plug-in & open & closed & closed & all & closed \\
\midrule
Rhythm alone & 69.0 & 63.0 & 62.5 & 76.2 & 79.0 \\
\midrule
CNNSpot & 86.4 \dsc{+3.5} & 76.9 \dsc{+5.3} & 80.3 \dsc{+2.3} & 69.8 \dsc{+16.3} & 55.3 \dsc{+19.8} \\
DeMamba & 85.7 \dsc{+4.5} & 82.1 \dsc{+3.6} & 75.2 \dsc{+3.6} & 74.5 \dsc{+13.1} & 74.0 \dsc{+11.6} \\
TimeSformer & 81.2 \dsc{+6.4} & 90.1 \dsc{$-$1.1} & 85.5 \dsc{$-$0.9} & 77.8 \dsc{+10.7} & 82.1 \dsc{+6.3} \\
Effort & 87.9 \dsc{+2.3} & 92.2 \dsc{$-$0.8} & 82.5 \dsc{+0.8} & 80.4 \dsc{+10.0} & 80.7 \dsc{+7.3} \\
\methodgf & 92.7 \dsc{+0.7} & 91.6 \dsc{$-$0.3} & 85.2 \dsc{+0.4} & 75.4 \dsc{+9.6} & 60.9 \dsc{+9.6} \\
\method & \best{93.8} \dsc{$-$1.2} & \best{94.6} \dsc{$-$1.8} & \best{88.3} \dsc{+0.1} & \best{85.2} \dsc{+6.7} & \best{83.5} \dsc{+4.4} \\
\bottomrule
\end{tabular}
\end{table}

\begin{table}[t]
\centering
\caption{\textbf{Ablation of the rhythm evidence} (AUROC, \%) on the scanned clips of \cref{tab:plugin}. Bold marks the best per block.}
\label{tab:abl-rhythm}
\scriptsize
\setlength{\tabcolsep}{5.2pt}
\begin{tabular}{@{}lcccc@{}}
\toprule
 & AIGVD & \multicolumn{2}{c}{DF26} & 2026 \\
 & closed & all & closed & closed \\
\midrule
\multicolumn{5}{@{}l}{\emph{Zero-shot score, no training}} \\
Reconstruction error $E_0$ & 49.5 & 41.8 & 35.5 & 38.0 \\
Periodicity, no frame baseline & \best{66.1} & 70.1 & 69.7 & 51.9 \\
Periodicity, keeping $j=0$ & 59.7 & 68.5 & 70.0 & 42.4 \\
Periodicity $\Ppk$ (ours) & 63.0 & \best{76.2} & \best{79.0} & \best{62.5} \\
\multicolumn{5}{@{}l}{\emph{Plug-in evidence, host \method}} \\
No plug-in & \best{96.4} & 78.5 & 79.0 & 88.2 \\
Reconstruction error & 95.5 & 76.1 & 76.3 & 85.8 \\
Scores of ReStraV & 93.5 & 74.8 & 74.1 & 84.6 \\
Periodicity, no positive part & 91.9 & \best{85.5} & \best{85.4} & 86.1 \\
Periodicity, positive part (ours) & 94.6 & 85.2 & 83.5 & \best{88.3} \\
\bottomrule
\end{tabular}
\end{table}

\begin{table}[t]
\centering
\caption{\textbf{Ablation of the detector} (AUROC, \%). Rows average 3 seeds, and the first row uses seed 0. Bold marks the best per block.}
\label{tab:abl-detector}
\scriptsize
\setlength{\tabcolsep}{3.2pt}
\begin{tabular}{@{}lccccc@{}}
\toprule
 & \multicolumn{2}{c}{AIGVD} & 2026 & \multicolumn{2}{c}{DF26} \\
 & open & closed & closed & all & closed \\
\midrule
\multicolumn{6}{@{}l}{\emph{Training data, one backbone}} \\
Benchmark data only & 68.9 & 81.2 & 79.3 & 62.1 & \best{65.2} \\
+ post-processed reals & 73.8 & 86.2 & 82.6 & 60.7 & 64.3 \\
+ \pairs, single head & \best{93.0} & \best{92.2} & \best{84.8} & \best{70.4} & 59.3 \\
Reals + \pairs only, \methodgf & 90.9 & 91.9 & \best{84.8} & 65.8 & 51.3 \\
\multicolumn{6}{@{}l}{\emph{Paths}} \\
Artifact path & \best{94.1} & 92.0 & \best{89.1} & 73.5 & 65.5 \\
Semantic path & 82.9 & 93.0 & 76.5 & 74.3 & \best{83.2} \\
Both paths, \method & 93.9 & \best{95.8} & 88.3 & \best{78.5} & 79.0 \\
\multicolumn{6}{@{}l}{\emph{Ensemble controls}} \\
Artifact path, two seeds & \best{94.5} & 92.5 & \best{89.5} & 73.9 & 66.0 \\
Semantic path, two seeds & 83.2 & 93.4 & 76.9 & \best{74.7} & \best{83.8} \\
Effort + TimeSformer & 84.3 & \best{93.6} & 82.7 & 70.5 & 73.7 \\
\multicolumn{6}{@{}l}{\emph{\pairs on another detector}} \\
Effort & 84.3 & \best{92.5} & 81.6 & \best{70.4} & \best{73.4} \\
Effort + \pairs & \best{93.1} & 77.3 & \best{85.0} & 68.4 & 47.6 \\
\bottomrule
\end{tabular}
\end{table}

}}
\method averages frame-level evidence over time, and most other detectors also do not measure temporal periodicity explicitly.
The plug-in therefore adds the training-free rhythm score $s_c$ to the score $s_d$ of any detector, including \method, without retraining it.
We use $s_c=\Ppk$ of \cref{sec:scan} because it assumes no period and thus also covers the period of five frames that a conversion from 24 to 30\,fps creates.
Let $z(\cdot)$ be a standardization computed on \emph{real} videos only.
The plug-in score is
\begin{equation}
s = z(s_d) + \max\big(0,\,z(s_c)\big).
\label{eq:plugin}
\end{equation}
After $z(\cdot)$ both scores measure how far a clip lies above typical real videos.
They can therefore be added without fitting any parameter on generated videos.
The rhythm score is continuous, so $z(s_c)$ subtracts its mean over real videos and divides by its standard deviation.
Detector scores are often saturated probabilities, and most real-video scores are below $10^{-3}$.
Inside the range of the real-video scores, $z(s_d)$ therefore maps a score to the standard-normal quantile of its mid-rank position among them.
Beyond that range it continues linearly with slope $1/\sigma$, where $\sigma$ is the standard deviation of the real-video scores.
The detector thus keeps its ordering of the clips that score above every real video.
A strong rhythm is evidence of generation.
A weak rhythm is not evidence of authenticity, because static scenes and post-processing can also weaken it.
The positive part therefore only adds evidence, and \cref{sec:exp-ablation} shows that it limits losses where the rhythm is weak.
For a clip whose rhythm is at or below the real mean, the fused score equals $z(s_d)$, so the detector decides alone.
We scan each clip at native resolution and downscale only short sides above 720 pixels, because rhythm strength depends on the scan scale.
Both standardizations use held-out real videos scanned in the same way, and all test sets share them.

\section{Experiments}
\label{sec:exp}

\subsection{Setup}
\label{sec:exp-setup}
\paragraph{Protocol.}
We follow the one-to-all protocol of AIGVDBench~\cite{aigvdbench}.
Detectors train on 14{,}000 Open-Sora~\cite{opensora} clips and 14{,}000 OpenVid~\cite{openvid} real clips and are tested on 20 open-source and 11 closed-source generators against 3{,}000 real clips.
We average the per-generator AUROC over open-source, unseen open-source and closed-source generators.
Without retraining, we also evaluate on DF26~\cite{df26}, which has ten generators released in 2025 and 2026 and permits content-paired 2AFC evaluation.
The 2026 models are single-shot clips of MiniMax~H3, Seedance~2.5~\cite{kivi} and Seedance~2.0~\cite{helioseedance}, compared against the DF26 reals.
We also test on ViF-Bench~\cite{skyra}, and \cref{sec:supp-testsets} in the supplement lists all test-set sizes.
All detectors use the same 8 frames from a shared $256^2$ cache, followed by method-specific input transforms.
No test clip enters detector training or the engine.
The engine uses the VAEs of Wan2.1, HunyuanVideo, CogVideoX and LTX-Video~\cite{wan,hunyuan,cogvideox,ltx}.
It turns 4{,}000 training reals into 4{,}818 training and 515 validation pairs.
\Cref{sec:supp-train,sec:supp-chain} in the supplement give further details.

\paragraph{Baselines.}
We retrain the 15 baselines in \cref{tab:main} under this protocol with their official implementations, including the 2026 method Native-Scale~\cite{nativescale}.
\Cref{sec:supp-baselines} in the supplement records our adaptations to sampling, numerical precision and training schedules, together with the original results reported by AIGVDBench.
Some reproductions remain below those results.

\subsection{Main results}
\label{sec:exp-main}
On AIGVDBench, \method reaches 93.9\% AUROC on open-source and 95.8\% on closed-source generators.
The strongest reproduced baselines there are ForgeLens and Effort, and \method exceeds them by 8.2\% and 3.3\%.
Without generator-produced training clips, \methodgf still reaches 90.9\% and 91.9\%.
This is well above the same backbone trained on Open-Sora without pairs, so the pairs carry most of the performance here.
\method also leads on the 2026 closed-source models and on all four DF26 averages.
Its DF26 all-generator AUROC of 78.5\% beats Effort by 8.1\%.
\Cref{tab:newgen} in the supplement reports each newer generator.
Over the four closed-source generators of DF26 and the three 2026 models, \method averages 83.0\% and exceeds the strongest baseline TimeSformer by 2.6\%.
Among the 2026 models, \method exceeds Effort by 17\% on Seedance~2.5.
On Kling~3.0 and on the Wan~2.2 and HunyuanVideo~1.5 generators of DF26, the training-free scan alone matches or exceeds every detector in \cref{tab:newgen}.
All methods perform poorly on ViF-Bench because its reals are re-encoded to 256p at 16\,fps.

\subsection{Plugging the rhythm into 19 detectors}
\label{sec:exp-plugin}
\Cref{tab:plugin} applies \cref{eq:plugin} to representative detectors, and \cref{tab:plugin-full} in the supplement covers all 19 rows of \cref{tab:main}.
The AIGVDBench columns use the 6{,}443 scanned clips.
Following \cref{sec:supp-comparable} in the supplement, their open-source average covers the 15 generators whose scans match the real clips, so the base scores differ slightly from \cref{tab:main}.

On DF26, all 19 detectors improve by 6.7\% to 23.7\%.
\Cref{fig:analysis}a shows where the gains come from.
Across 39 generators, the gain averaged over detectors follows the zero-shot AUROC of the rhythm with $r=0.89$ and turns positive once that AUROC passes about 55\%.
The plug-in therefore helps where a rhythm is measurable and changes little elsewhere.
This explains the smaller effect on AIGVDBench's closed-source generators and the 2026 models, where the rhythm is weaker.
On these two groups, 13 and 18 of the 19 detectors still improve.
\Cref{fig:analysis}b shows that weaker detectors gain more, with $r=-0.95$, although even \method gains.
Detectors trained with VAE pairs also gain on DF26, so the temporal scan adds evidence that frame-level training misses.
On Grok Imagine, Kling~3.0 and Wan~2.6, the plug-in adds 5\% to 7\% to \method, as \cref{tab:newgen} in the supplement shows.
With the plug-in, \method remains the strongest of the 19 detectors on AIGVDBench, DF26 and the 2026 models, and it reaches 85.2\% on DF26.

\subsection{Ablations}
\label{sec:exp-ablation}
\Cref{tab:abl-rhythm,tab:abl-detector} separate the contributions of the rhythm evidence, the \pairs and the two paths.
\Cref{tab:ablation} in the supplement adds leave-one-VAE-out and other variants.

\paragraph{Rhythm evidence.}
The signal comes from periodicity, not from the size of the reconstruction error.
If lower error is treated as generated, the error falls below chance on DF26, whereas periodicity is 34.4\% higher.
Excluding the first window frame and subtracting the frame baseline both help on DF26 and the 2026 models.
As plug-in evidence, only periodicity helps.
The reconstruction error and the scores of another weak detector both lower \method on DF26, whereas periodicity raises it.
The positive part limits losses on AIGVDBench and the 2026 models, where the rhythm is weak.

\paragraph{VAE pairs.}
On one backbone, post-processed real clips alone help on closed-source generators, which suggests less reliance on processing artifacts.
Adding the VAE reconstructions raises the open-source AUROC by 19.2\%.
Leaving out any one of the four VAEs keeps it between 89.1\% and 93.4\%, so the engine does not rely on a single VAE.
On DF26's closed-source generators, however, the pairs lower both this backbone and Effort.
The semantic path therefore omits them, so that \method keeps both strengths.

\paragraph{Two paths.}
\Cref{fig:analysis}c shows that the two paths succeed on different generators.
On the 43 unseen generators, their AUROCs barely correlate, with $r=0.30$.
The artifact path is better on 19 of the 25 open-source generators, which fits the pairs teaching decoding artifacts.
The semantic path is better on all four closed-source models of DF26.
Their sum beats the better path on 21 of 43 generators and is never below their average.
This is not a generic ensemble effect.
The sum of the two best baselines trails \method in every column, and two seeds of one path keep its weakness.

\ifrobust
\begin{figure}[t]
\centering
\includegraphics[width=\columnwidth]{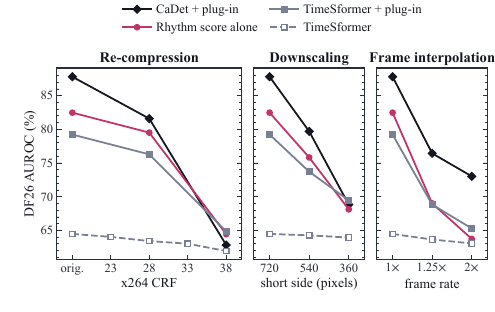}
\caption{\textbf{Robustness to post-processing} on 250 DF26 clips (AUROC, \%). Real and generated clips receive the same processing, and every calibration stays fixed.}
\label{fig:robustness}
\end{figure}

\paragraph{Robustness to post-processing.}
\Cref{fig:robustness} applies three common kinds of processing to real and generated DF26 clips alike.
We sample 100 real clips and 15 clips of each DF26 generator and keep their first 2.5\,s.
Re-compression uses x264 at CRF 23 to 38.
Downscaling reduces the short side from 720 to 540 or 360 pixels, and interpolation raises the frame rate by 1.25 or 2 times.
The rhythm score changes little at CRF 28 and stays above chance after downscaling and frame interpolation.
The plug-in therefore keeps adding evidence after processing, and it raises TimeSformer under every setting.
With the plug-in, \method remains the strongest detector under frame interpolation and under moderate compression or downscaling.
Under the strongest settings, it is on par with TimeSformer with the plug-in.
\Cref{sec:supp-robust} in the supplement gives the settings.
\fi

\section{Conclusion and Limitations}
\label{sec:conclusion}
We identify \rhythm as a hidden temporal signature of frame grouping in 3D causal video VAEs.
Across 39 scannable generator variants, a training-free phase scan finds it in open- and closed-source families, and controls and a scale law support its origin in decoding.
Pairs built from real videos enable generator-free detection, a semantic path makes our detector the strongest in our comparison, and the plug-in improves every tested detector on DF26.
Because the rhythm comes from decoding rather than from content, it complements detectors that learn appearance, and the plug-in adds it to them without any training.

Like other pixel-level traces, the rhythm needs some motion and a scan scale close to the generation scale, and strong re-encoding or frame-rate conversion can weaken it.
Future generators may also change their VAEs.
The engine needs only real videos and a VAE, so a new VAE can be added to training without collecting any generated video.

{\small
\bibliographystyle{plainnat}
\bibliography{main}
}
\twocolumn[{\centering\Large\bfseries Supplementary Material\par\vspace{1.4em}}]
\renewcommand{\theHsection}{S\arabic{section}}
\renewcommand{\theHfigure}{S\arabic{figure}}
\renewcommand{\theHtable}{S\arabic{table}}
\setcounter{section}{0}\renewcommand{\thesection}{S\arabic{section}}
\setcounter{figure}{0}\renewcommand{\thefigure}{S\arabic{figure}}
\setcounter{table}{0}\renewcommand{\thetable}{S\arabic{table}}

\section{Additional analysis of VAE Rhythm}
\label{sec:supp-analysis}
\begin{figure*}[tp]
\centering
\includegraphics[width=\textwidth]{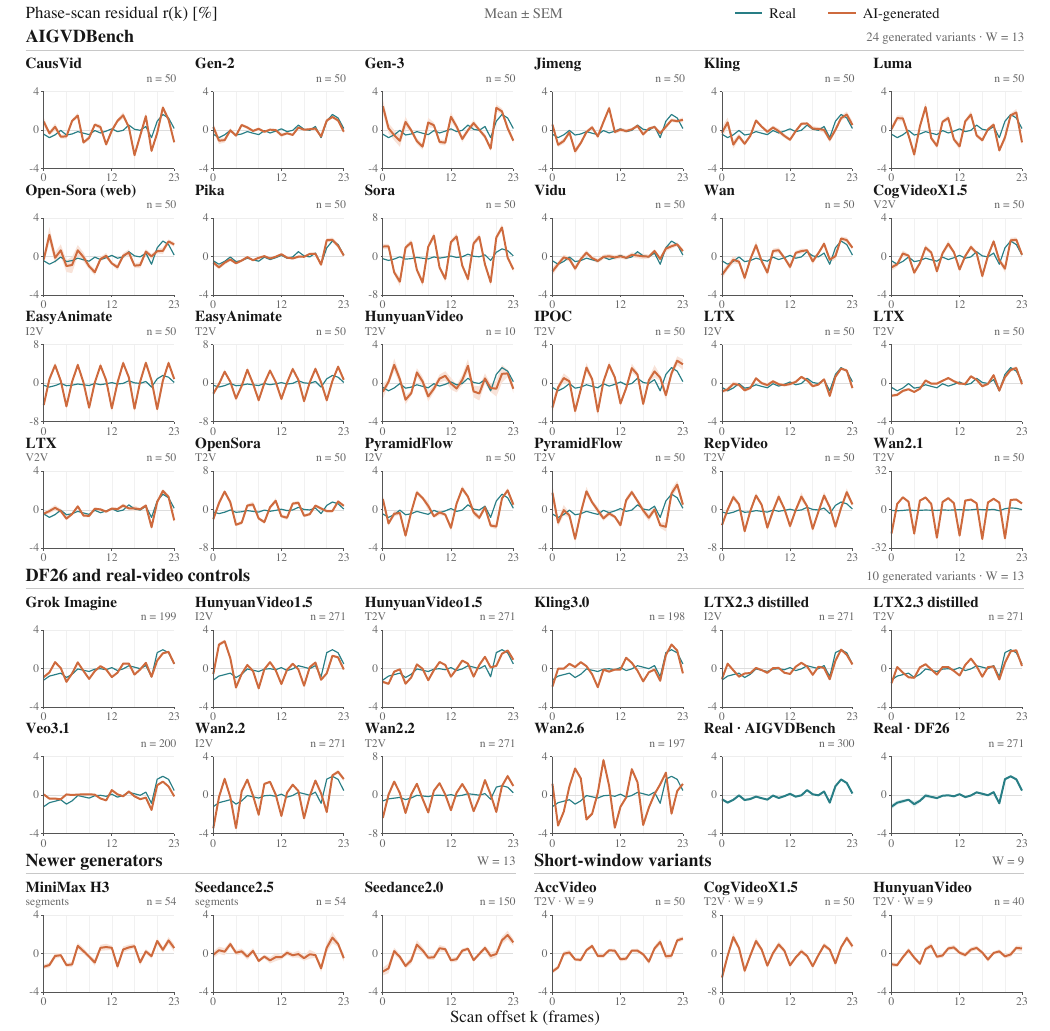}
\caption{\textbf{VAE Rhythm across all 39 generator variants.} Mean residuals $r(k)$ with one standard error. Orange curves are generated clips, teal curves are real clips of the same benchmark, and $n$ is the number of clips or source segments.}
\label{fig:rhythm-curves-full}
\end{figure*}

This section collects the analyses that \cref{sec:evidence} summarizes.
They cover the full atlas of all 39 variants in \cref{fig:rhythm-curves-full}, per-variant statistics, controls, the zero-shot census, phase and spectral summaries, family-dependent periods, the scale law and the choice of measurement VAE.

\Cref{fig:rhythm-curves-full} plots every available 24-offset profile in the census subsets at native resolution capped at 480p.
The profiles have no phase alignment, smoothing, per-video amplitude normalization or real-reference subtraction.
Vertical ranges start at $\pm4\%$ and expand by powers of two as needed, and each panel labels its limits.
The bottom row has no real overlays, because the newer collections lack their own real cohort and the short-window scans lack a matched $W=9$ reference.
HunyuanVideo appears under both window settings.
The five 16-frame 2D-VAE variants cannot support 24 offsets and are omitted.

\subsection{Per-variant statistics}
\label{sec:supp-stats}
\begin{table}[t]
\centering
\caption{\textbf{Real vs.\ generated statistics of \rhythm} for the 36 window-matched variants. \cref{sec:supp-stats} defines the columns.}
\label{tab:rhythm-stats}
\scriptsize
\setlength{\tabcolsep}{3.2pt}
\begin{tabular}{@{}lrlccccc@{}}
\toprule
variant & $n$ & stat. & median & AUROC & $p$ & above & $R$ \\
\midrule
\multicolumn{8}{@{}l}{\emph{AIGVDBench, open-source}} \\
Wan2.1 T2V & 50 & $\Cfour$ & 0.96 / 0.11 & 99.5 & $<$.001 & 96 & 0.99 \\
EasyAnimate I2V & 50 & $\Cfour$ & 0.81 / 0.11 & 94.2 & $<$.001 & 82 & 0.86 \\
EasyAnimate T2V & 50 & $\Cfour$ & 0.74 / 0.11 & 89.8 & $<$.001 & 80 & 0.59 \\
RepVideo & 50 & $\Cfour$ & 0.61 / 0.11 & 89.3 & $<$.001 & 70 & 0.71 \\
CogVideoX1.5 V2V & 50 & $\Cfour$ & 0.43 / 0.11 & 77.4 & $<$.001 & 50 & 0.49 \\
Open-Sora & 50 & $\Cfour$ & 0.36 / 0.11 & 77.0 & $<$.001 & 32 & 0.64 \\
IPOC & 50 & $\Cfour$ & 0.37 / 0.11 & 76.1 & $<$.001 & 40 & 0.67 \\
Pyramid-Flow I2V & 50 & $\Cfour$ & 0.26 / 0.11 & 71.3 & $<$.001 & 38 & 0.37 \\
Pyramid-Flow T2V & 50 & $\Cfour$ & 0.20 / 0.11 & 71.0 & $<$.001 & 24 & 0.33 \\
LTX V2V & 50 & $\Cfour$ & 0.26 / 0.11 & 65.5 & .002 & 14 & 0.43 \\
LTX I2V & 50 & $\Cfour$ & 0.16 / 0.11 & 58.9 & .107 & 10 & 0.40 \\
LTX T2V & 50 & $\Cfour$ & 0.13 / 0.11 & 52.0 & .978 & 10 & 0.29 \\
\multicolumn{8}{@{}l}{\emph{AIGVDBench, closed-source}} \\
Sora & 50 & $\Cfour$ & 0.78 / 0.11 & 90.4 & $<$.001 & 72 & 0.70 \\
CausVid & 50 & $\Cfour$ & 0.35 / 0.11 & 80.0 & $<$.001 & 36 & 0.62 \\
Luma & 50 & $\Cfour$ & 0.49 / 0.11 & 79.3 & $<$.001 & 52 & 0.76 \\
Gen-3 & 50 & $\Cfour$ & 0.39 / 0.11 & 78.6 & $<$.001 & 36 & 0.68 \\
Vidu & 50 & $\Cfour$ & 0.34 / 0.11 & 68.6 & $<$.001 & 42 & 0.21 \\
Open-Sora (web) & 50 & $\Cfour$ & 0.16 / 0.11 & 62.9 & .016 & 24 & 0.31 \\
Kling & 50 & $\Cfour$ & 0.19 / 0.11 & 61.8 & .026 & 16 & 0.44 \\
Jimeng & 50 & $\Cfour$ & 0.19 / 0.11 & 59.4 & .097 & 14 & 0.56 \\
Wan (web) & 50 & $\Cfour$ & 0.20 / 0.11 & 58.7 & .107 & 32 & 0.44 \\
Pika & 50 & $\Cfour$ & 0.06 / 0.11 & 35.0 & 1.0 & 6 & 0.32 \\
Gen-2 & 50 & $\Cfour$ & 0.05 / 0.11 & 30.7 & 1.0 & 0 & 0.22 \\
\multicolumn{8}{@{}l}{\emph{DF26}} \\
Wan~2.2 T2V & 271 & $\Cfour$ & 0.66 / 0.11 & 94.7 & $<$.001 & 84 & 0.92 \\
Wan~2.2 I2V & 271 & $\Cfour$ & 0.46 / 0.11 & 88.4 & $<$.001 & 62 & 0.88 \\
Wan~2.6 & 197 & $\Ppk$ & 0.49 / 0.24 & 87.7 & $<$.001 & 54 & 0.94 \\
Kling~3.0 & 198 & $\Ceight$ & 0.50 / 0.26 & 87.2 & $<$.001 & 50 & 0.75 \\
HunyuanVideo~1.5 T2V & 271 & $\Cfour$ & 0.29 / 0.11 & 76.7 & $<$.001 & 41 & 0.74 \\
LTX~2.3 T2V & 271 & $\Ceight$ & 0.42 / 0.26 & 76.5 & $<$.001 & 35 & 0.48 \\
HunyuanVideo~1.5 I2V & 271 & $\Cfour$ & 0.24 / 0.11 & 73.6 & $<$.001 & 34 & 0.80 \\
Grok Imagine & 199 & $\Cfour$ & 0.22 / 0.11 & 70.0 & $<$.001 & 24 & 0.64 \\
LTX~2.3 I2V & 271 & $\Cfour$ & 0.15 / 0.11 & 60.6 & $<$.001 & 14 & 0.49 \\
Veo~3.1 & 200 & $\Cfour$ & 0.14 / 0.11 & 57.9 & .016 & 14 & 0.42 \\
\multicolumn{8}{@{}l}{\emph{2026 models}} \\
Seedance~2.0 & 150 & $\Cfour$ & 0.24 / 0.11 & 73.5 & $<$.001 & 27 & 0.45 \\
MiniMax~H3 & 54 & $\Cfour$ & 0.20 / 0.11 & 67.3 & $<$.001 & 26 & 0.41 \\
Seedance~2.5 & 54 & $\Cfour$ & 0.19 / 0.11 & 66.5 & $<$.001 & 22 & 0.26 \\
\midrule
\multicolumn{8}{@{}l}{\emph{Real cohorts, period 4}} \\
AIGVDBench & 300 & $\Cfour$ & 0.11 & -- & -- & 5 & 0.08 \\
DF26 & 271 & $\Cfour$ & 0.11 & -- & -- & 5 & 0.07 \\
\bottomrule
\end{tabular}
\end{table}

\Cref{tab:rhythm-stats} lists the values behind \cref{fig:rhythm-stats}.
The median column gives the statistic of the generated clips and of the real clips of the same benchmark.
$p$ comes from a one-sided Mann--Whitney test with Holm correction over the 36 variants.
The above column is the share of generated clips above the 95th percentile of real clips in percent, which is 5\% for real clips by definition.
$R$ is the phase concentration at the period of the variant.
Each variant is compared with the real clips of its own benchmark, and the 2026 models are compared with the DF26 reals.
The statistic is the one of \cref{tab:census}, which is $\Cfour$ by default, $\Ceight$ for Kling~3.0 and LTX~2.3 T2V, and $\Ppk$ for Wan~2.6.
We chose these three exceptions from the curves in \cref{fig:rhythm-curves-full}.
$R$ uses all clips of a variant at the period of its statistic, which is 5 for Wan~2.6.
We compute it after subtracting the mean profile of the 300 AIGVDBench reals as in \cref{sec:supp-scan} and test it with the Rayleigh test.
At period 8, the real cohorts reach $R=0.07$ on AIGVDBench and $R=0.16$ on DF26.
These values differ slightly from the histograms in \cref{fig:supp-phaselock}, which sample up to 100 clips per generator.
The three short-window variants are excluded because they lack a matched real reference.

\subsection{Controls}
\label{sec:controls}
Four controls support the decoding origin.
\emph{VAE reconstruction.} Real clips passed through Wan2.1 acquire a period-4 signature.
$\Cfour$ separates the reconstructions from their originals with 90\% AUROC, and the phase is aligned in all 163 clips.
\emph{Temporal shift.} Removing two leading frames shifts the measured phase by two positions, as the lattice predicts.
A statistic fixed to phase 0 falls to 21\% AUROC, while its phase-independent counterpart stays at 67\%.
\emph{Codec.} Some generated videos use an IBBBP coding pattern, and its four-frame interval could confound the VAE signal.
However, 120 DF26 real clips re-encoded with Grok's x264 parameters remain indistinguishable from their originals under the scan, at about 49.0\% AUROC, and their phases stay dispersed.
Re-encoding 300 AIGVDBench reals as IBBBP gives 50.2\%.
HEVC B-pyramids in GenBuster reals produce a weak period-4 component at a different phase, and the 24-to-16\,fps conversion of ViF-Bench produces period 2.
\emph{Motion.} The rhythm is weak in nearly static clips, and its measured strength increases with motion.
These controls rule out the tested codec configurations as the main explanation.

\subsection{Zero-shot census protocol and results}
\label{sec:supp-census}
\begin{table*}[t]
\centering
\caption{\textbf{Zero-shot AUROC (\%) per generator} using the rhythm statistic alone without training. The AIGVDBench 720p column uses smaller subsets, and $^\ddagger$marks short-window scans without a matched real reference.}
\label{tab:census}
\scriptsize
\setlength{\tabcolsep}{3.2pt}
\resizebox{\textwidth}{!}{%
\begin{tabular}{@{}llr@{\hskip 9pt}llrr@{\hskip 9pt}llrr@{}}
\toprule
\multicolumn{3}{@{}l}{\textbf{AIGVDBench, open-source}} & \multicolumn{4}{l}{\textbf{AIGVDBench, closed-source}} & \multicolumn{4}{l@{}}{\textbf{DF26 and 2026 models}} \\
generator & VAE & 480p & generator & VAE & 480p & 720p & generator & VAE & 480p & 720p \\
\midrule
Wan2.1 T2V & Wan 4$\times$ & 99.5 & Sora & Unknown & 90.4 & 84 & Wan~2.2 T2V / I2V & Wan 4$\times$ & 94.7 / 88.4 & 98.8 / 96.7 \\
EasyAnimate I2V / T2V & own 4$\times$ & 94.2 / 89.8 & Luma & Unknown & 79.3 & 89 & HunyuanVideo~1.5 T2V / I2V & own 4$\times$ & 76.7 / 73.6 & 84.3 / 81.9 \\
RepVideo & CogX 4$\times$ & 89.3 & Gen-3 & Unknown & 78.6 & 80 & LTX~2.3 T2V / I2V & own 8$\times$ & 76.5 / 60.6 & 74.3 / 60.0 \\
CogVideoX1.5 T2V / V2V & CogX 4$\times$ & 77.8$^\ddagger$ / 77.4 & CausVid & Wan 4$\times$ & 80.0 & 77 & Kling~3.0 & Unknown, 8-frame & 87.2 & 88.2 \\
Open-Sora & own 4$\times$ & 77.0 & Wan (web) & Wan 4$\times$ & 58.7 & 78 & Wan~2.6, 30\,fps & Unknown, period 5 & 87.7 & 96.0 \\
IPOC & CogX? 4$\times$ & 76.1 & Vidu & Unknown & 68.6 & 73 & Grok Imagine & Unknown & 70.0 & 79.4 \\
Pyramid-Flow I2V / T2V & own 8$\times$ & 71.3 / 71.0 & Kling & Unknown & 61.8 & 68 & Veo~3.1 & Unknown & 57.9 & 73.4 \\
HunyuanVideo T2V & own 4$\times$ & 64.6$^\ddagger$ & Open-Sora (web) & Unknown & 62.9 & 59 & MiniMax~H3, 2026 & own 4$\times$ & 67.3 & 78.9 \\
AccVideo & Hunyuan 4$\times$ & 61.4$^\ddagger$ & Jimeng & Unknown & 59.4 & 58 & Seedance~2.5, 2026 & Unknown & 66.5 & 67.6 \\
LTX T2V / I2V / V2V & own 8$\times$ & 52.0 / 58.9 / 65.5 & Pika & Unknown & 35.0 & 23 & Seedance~2.0, 2026 & Unknown & 73.5 & 78.5 \\
Five 2D-VAE variants & SD 2D & n/a & Gen-2 & Unknown & 30.7 & 26 & DF26 reals & -- & 50 & 50 \\
\bottomrule
\end{tabular}}
\end{table*}

\Cref{tab:census} scans each clip once with the frozen Wan2.1 VAE and scores it with the fraction of residual variance at the period of its family.
The statistic is $\Cfour$ by default, $\Ceight$ for the eight-frame families Kling~3.0 and LTX~2.3 T2V, and $\Ppk$ for the period-5 Wan~2.6.
The AUROC measures how well this score separates the clips of a generator from the real clips of the same benchmark, with generated clips as positives.
An AUROC of 50\% is chance, and values below 50\% mean that the real clips are scored as more periodic than the generated ones.
For AIGVDBench, we score 50 clips per generator against 300 reals.
For DF26, we score all 2{,}420 generated clips against the 271 DF26 reals.
For the 2026 models, we score 54 MiniMax~H3, 54 Seedance~2.5 and 150 Seedance~2.0 single-generation clips against the same reals.
Scans use native resolution capped at 480p.
For AIGVDBench, the 720p column uses a smaller subset of 25 clips per closed-source generator.
For DF26 and the 2026 models, the 720p column rescans the same clips as the 480p column at native resolution capped at 720p.
Slash-separated AUROCs follow the task order in the generator column.
Short-window results compare $W=9$ scans with $W=13$ reals and are not matched-window evidence.
Unknown denotes an unverified source VAE, and the listed periods are observed rather than confirmed compression factors.
The CogVideoX attribution of IPOC is unverified.
CausVid and Wan (web) are Wan2.1-based models that AIGVDBench lists as closed-source.
The five 16-frame 2D-VAE variants are too short for 24 offsets.
The per-family statistic follows the periods observed in \cref{fig:rhythm-curves-full}, whereas the plug-in of \cref{sec:plugin} uses the single statistic $\Ppk$ for all generators.

Open-source $4\times$-VAE generators are separated best, and Wan2.1 is close to perfect.
The $8\times$ families are weaker under the $4\times$ measurement VAE, and the three short-window entries are marked as unmatched comparisons.
Most closed-source generators with unknown VAEs and generation resolutions are also separated from the reals, and 720p scans raise Luma, Wan (web) and Vidu.
Gen-2 and Pika remain below chance at both scales, which is consistent with their curves.
On DF26, generators released in 2025 and 2026 are separated from the reals of the same benchmark.
At native 720p, every DF26 and 2026 generator except the two LTX~2.3 variants scores higher, and Veo~3.1 gains the most.
All four closed-source DF26 generators therefore show measurable rhythm once the scan matches their delivery resolution.
Excluding the three short-window entries, 34 of the remaining 36 variants exceed 50\% at the default 480p scale, 22 reach at least 70\% and 10 reach at least 80\%.

\subsection{Phase, spectrum and barcode summaries}
\label{sec:supp-figs}
\begin{figure*}[t]
\centering
\includegraphics[width=\textwidth]{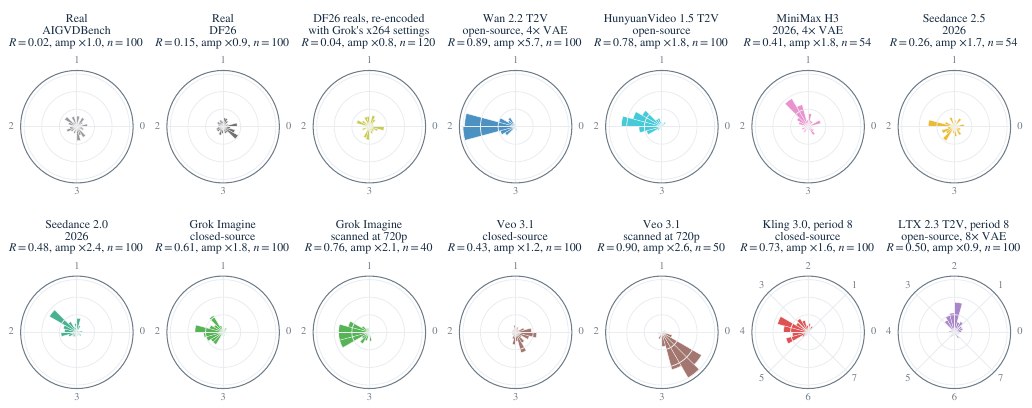}
\caption{\textbf{Phase locking.} Phase histograms of the period-4 component, with the concentration $R$ and the amplitude ratio of each group. The indicated groups use period 8.}
\label{fig:supp-phaselock}
\end{figure*}

\begin{figure}[tb]
\centering
\includegraphics[width=\linewidth]{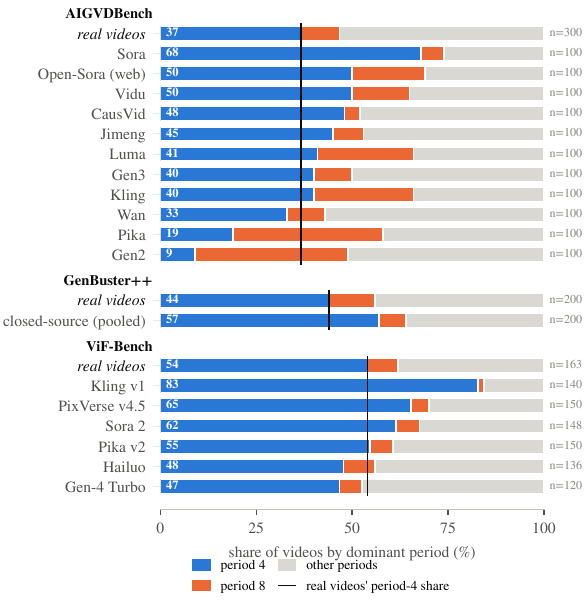}
\caption{\textbf{Dominant period} of the residual curve. Bars show the share of clips whose strongest period is 4 in blue, 8 in orange or another period in gray. The black line marks the real period-4 share of the same benchmark.}
\label{fig:supp-period}
\end{figure}

\Cref{fig:supp-phaselock} includes HunyuanVideo~1.5, LTX~2.3, Wan~2.6, Seedance~2.5 and the 480p and 720p scans of Grok and Veo~3.1.
It typically samples up to 100 clips per generator.
The codec controls use 120 real clips, and the amplitude ratio uses the median amplitude relative to AIGVDBench reals.
These summaries use sampled clips and the real-reference correction.
They complement the raw profiles of the full collections in \cref{fig:rhythm-curves-full}, and their sample counts and processing differ.
\Cref{fig:supp-barcode} stacks the corrected residual curves by generator, with one clip per row sorted by periodicity and up to 100 clips per group.
The displayed latent generators show aligned bands at period 4.
The bands are at period 8 for Kling~3.0 and LTX~2.3 and at period 5 for Wan~2.6, whereas real clips lack a comparable pattern.
\Cref{fig:supp-period} summarizes the dominant periods from a 256p scan with $K{=}60$ of closed-source clips in AIGVDBench, GenBuster and ViF-Bench.
The period-4 share is 68\% for Sora against 37\% for real clips, and 83\% for ViF kling-v1 against 54\%.

\begin{figure}[tb]
\centering
\includegraphics[width=\linewidth]{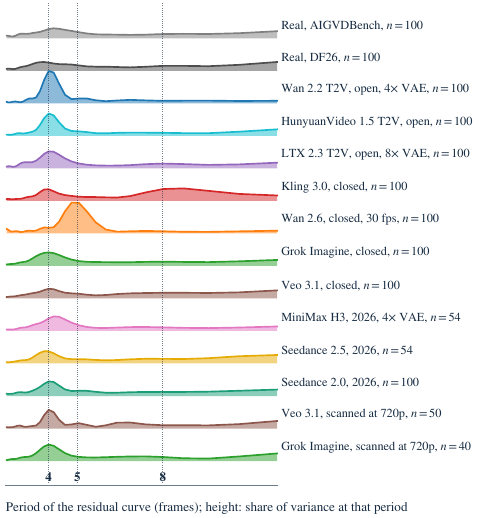}
\caption{\textbf{Spectral summaries.} Mean normalized periodograms of the sampled profiles. Sample counts are beside each group.}
\label{fig:supp-spectrum}
\end{figure}

\begin{figure*}[t]
\centering
\includegraphics[width=0.86\textwidth]{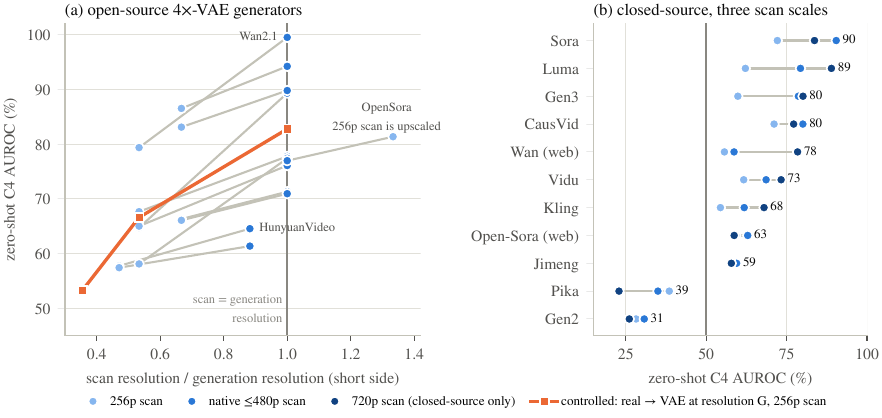}
\caption{\textbf{The scale law.} (a)~Zero-shot $\Cfour$ AUROC of 12 open-source $4\times$-VAE generators at 256p in light blue and at native resolution capped at 480p in dark blue. Orange shows the controlled real-through-VAE experiment. (b)~Closed-source models at three scan scales, labelled with their maximum AUROC across scales.}
\label{fig:scale}
\end{figure*}

\subsection{Family-dependent periods}
\label{sec:supp-periods}
Four-frame oscillations are prominent in several $4\times$-VAE families in \cref{fig:rhythm-curves-full}.
The curves also contain longer-period structure, as in Kling~3.0 and Pyramid-Flow.
LTX can contain both an eight-frame component and its four-frame harmonic.
Wan~2.6 instead shows a pattern of about five frames, which is consistent with a change from 24 to 30\,fps delivery.
These differences motivate the period-independent statistic $\Ppk$ instead of a test restricted to period 4.
The phase distributions and spectra in \cref{fig:supp-phaselock,fig:supp-spectrum} provide complementary summaries.
Unlike the raw profiles in \cref{fig:rhythm-curves-full}, they use the real-reference correction described in \cref{sec:supp-scan}.

\subsection{The scale law}
\label{sec:supp-scale}
As \cref{fig:scale} shows, rhythm strength depends strongly on the ratio $\rho=R_{\mathrm{scan}}/G$ of scan resolution to generation resolution.
We reconstruct real clips at short side $G$ and scan them at 256p.
Their median $\Cfour$ falls from 0.64 at $\rho=1$ to 0.33 at $\rho=0.53$ and approaches the real-video level of 0.18 at $\rho=0.36$.
Upscaling before scanning does not recover information lost at generation, and cropping is not equivalent to changing this ratio.
Native-resolution scans capped at 480p improve the zero-shot AUROC over 256p scans.
Sora rises from 72\% to 90\%, Luma from 62\% to 79\% and Wan2.1 from 79\% to 99.5\%.
Scanning at 720p raises most DF26 and 2026 generators further, as \cref{tab:census} lists.
The same scale dependence motivates the range of reconstruction resolutions that the engine uses in \cref{sec:engine}.

\subsection{Measurement VAEs}
\label{sec:supp-rulers}
The measurement VAE need not match the source generator.
The Wan2.1 VAE detects rhythm in videos from CogVideoX, HunyuanVideo, Pyramid-Flow, Grok and Seedance, and the CogVideoX VAE gives comparable results.
The HunyuanVideo and LTX VAEs transfer less well, so we measure with the Wan2.1 VAE.
On Kling~3.0, it gives 87\% AUROC, against 57\% with the $8\times$ LTX VAE.

\section{Scan settings and budget}
\begin{figure}[!t]
\centering
\includegraphics[width=\columnwidth]{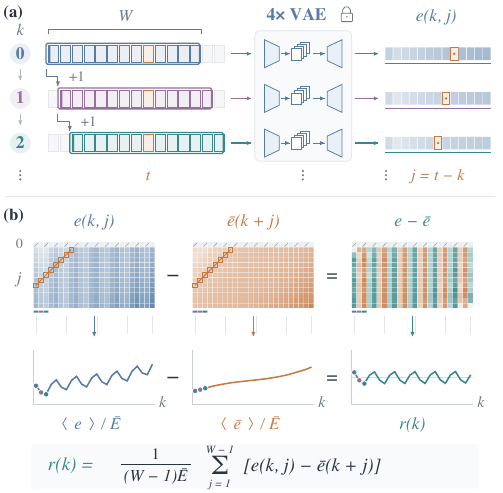}
\caption{\textbf{Phase scan in detail.} This figure expands \cref{fig:teaser}b. (a)~A $W$-frame window shifts by one frame per pass through the same frozen VAE. (b)~Subtracting the same-frame mean along the orange diagonal from the stacked errors and averaging over $j>0$ gives $r(k)$. Values are schematic.}
\label{fig:scan}
\end{figure}

\label{sec:supp-scan}
Unless stated otherwise, scans use the Wan2.1 VAE in bf16, $K{=}24$ offsets and $W{=}13$-frame windows that span the first $K+W-1=36$ frames.
Before the scan, every clip is re-encoded with x264 CRF~23 at native resolution capped at 480p, or at 720p where specified.
The plug-in evaluations in \cref{tab:plugin,tab:plugin-full,tab:newgen} cap the native resolution at 720p instead, for test clips and calibration reals alike.
The 33-frame AIGVDBench clips use $K=24$ and $W=9$.
These are all 50 AccVideo clips, all 50 CogVideoX1.5-T2V clips and 40 of the 50 HunyuanVideo-T2V clips, and the remaining ten HunyuanVideo clips use $W=13$.
\Cref{fig:rhythm-curves-full} separates these window settings and shows the original normalized profiles without real-reference subtraction.
Matched $W=9$ real scans at native 480p are unavailable, so the affected census AUROCs are marked and not counted as matched-window evidence.
The dominant-period analysis in \cref{fig:supp-period} instead uses longer clips and $K{=}60$ offsets, because 60 is divisible by 2, 3, 4, 5, 6, 10 and 12.
A 480p scan takes about 13\,s per clip on an H200 and 52\,s on an A6000.
A 720p scan takes about 30\,s on an H200 and 2\,min on an A6000.

\paragraph{Comparable scans.}
\label{sec:supp-comparable}
The five 16-frame 2D-VAE generators of AIGVDBench can only be scanned with $K{=}8$ and $W{=}9$, while the real clips use $K{=}24$ and $W{=}13$.
All zero-shot and plug-in averages over AIGVDBench's open-source generators therefore cover the other 15 generators.
\Cref{tab:main} keeps all 20, because detector scores do not involve the scan.

\begin{figure*}[t]
\centering
\includegraphics[width=\textwidth]{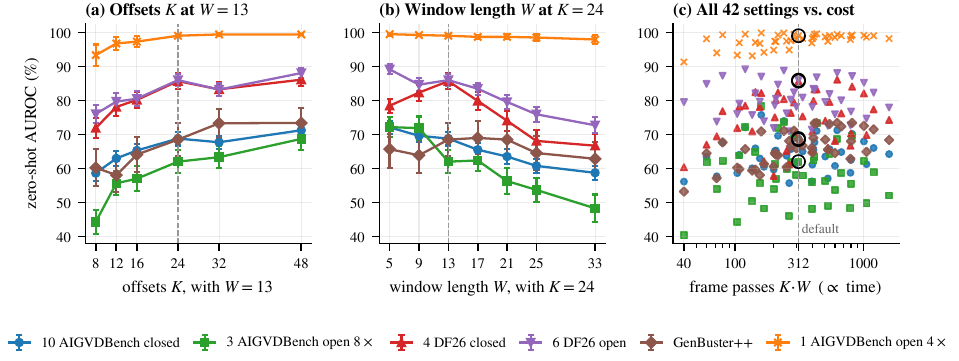}
\caption{\textbf{Zero-shot AUROC versus scan budget.} Points are group means with bootstrap standard errors. (a)~Offsets $K$ at $W{=}13$. (b)~Window length $W$ at $K{=}24$. (c)~All 42 settings against their cost. The default setting is circled.}
\label{fig:scan-budget}
\end{figure*}

\paragraph{Scan budget.}
The cost of a scan is $K\times W$ VAE frame passes.
On an H200 at $1280\times720$ in bf16 with one window per batch, the per-clip wall time is $0.25\,\mathrm{s}+94.9\,\mathrm{ms}\times KW$, with $R^2=1.000$ over 16 settings.
This gives 3.9\,s for $8\times5$, 29.9\,s for the default $24\times13$ and 150\,s for $48\times33$.
Decoding adds less than 0.1\,s.
Batching 2 or 4 windows keeps the time at 30.4\,s but raises memory from 10.7 to 20.8 and 40.9\,GB.
\Cref{tab:scan-budget,fig:scan-budget} vary $K$ and $W$ on a 1{,}290-clip subset of the census with at least 80 frames per clip.
These scans use native resolution capped at 720p and the census statistic.
The subset contains ten AIGVDBench closed-source generators, the $4\times$ Wan2.1-T2V, the $8\times$ Pyramid-Flow T2V/I2V and LTX-T2V, the ten DF26 generators and the 13 generators of GenBuster++~\cite{busterxpp}.
It uses 30 clips per generator and 300, 150 and 60 reals.
Time is the measured per-clip wall time on an H200 at $1280\times720$ in bf16 with batch size 1, or the fit above, which reproduces the measured settings within 0.3\,s.
At the default setting, the bootstrap standard errors of the group means are 2.0\%, 0.4\%, 3.4\%, 2.4\%, 1.7\% and 4.8\% in the column order of \cref{tab:scan-budget}.
One $48\times33$ scan per clip yields every setting by truncating the error table to $E[{:}K,{:}W]$.
This works because the Wan2.1 VAE is temporally causal, so the reconstruction of the first $W'$ frames of a window does not depend on later frames.
On the timing clips, truncated tables are identical to direct scans at the same batch size.
The group AUROCs at $24\times13$ agree with the earlier batch-2 scans within 0.2\%.
Windows longer than 13 frames hurt monotonically.
At $K{=}24$, a window of $W{=}33$ loses up to 19\% per group, which is consistent with the rhythm being concentrated in the first frames after each window start.
Shorter windows are competitive on average but depend on the generator.
A window of $W{=}5$ gains 9\% to 18\% on OpenSora-web, Wan (web), Jimeng, LTX-T2V and Pyramid-Flow, and it loses 10\% to 33\% on Veo~3.1, CausVid and Gen-3.
A window of $W{=}13$ is the best for the DF26 closed-source group.
More offsets help monotonically.
At $W{=}13$, $K{=}48$ adds up to 6.7\% over $K{=}24$ at twice the cost, while $K{\le}16$ loses up to 18\%.
The $32\times5$ setting spends the budget on offsets instead of window length at half the cost of the default.
It improves five of the six groups by up to 16\% and lowers the DF26 closed-source group by 3.5\%.
We keep $24\times13$ throughout rather than select a setting on these test clips.
Additional budget is better spent on offsets than on longer windows.

\begin{table*}[t]
\centering
\caption{\textbf{Scan budget} (zero-shot AUROC, \%). The table gives accuracy and per-clip time for different numbers of offsets $K$ and window lengths $W$. Bold marks the best per column, and $^{*}$ marks a time from the fit.}
\label{tab:scan-budget}
\small
\setlength{\tabcolsep}{4.5pt}
\begin{tabular}{@{}cccc|ccc|cc|c@{}}
\toprule
 & & & time & \multicolumn{3}{c|}{AIGVDBench} & \multicolumn{2}{c|}{DF26} & \\
$K$ & $W$ & $KW$ & (s) & 10 closed & 1 open $4\times$ & 3 open $8\times$ & 4 closed & 6 open & GB++ \\
\midrule
\multicolumn{10}{@{}l}{\emph{Window length $W$, with $K{=}24$}} \\
24 & 5 & 120 & 11.6 & 72.1 & 99.6 & 72.2 & 78.5 & \best{89.2} & 65.7 \\
24 & 9 & 216 & 20.7 & 69.7 & 99.3 & 71.9 & 82.3 & 84.7 & 63.9 \\
\textbf{24} & \textbf{13} & 312 & 29.9 & 68.8 & 99.1 & 62.0 & 85.7 & 86.0 & 68.5 \\
24 & 17 & 408 & 39.0 & 65.6 & 98.8 & 62.4 & 79.9 & 83.5 & 69.0 \\
24 & 21 & 504 & 48.1 & 63.5 & 98.8 & 56.3 & 74.1 & 79.6 & 68.5 \\
24 & 25 & 600 & 57.3 & 60.8 & 98.6 & 53.7 & 68.2 & 76.0 & 64.6 \\
24 & 33 & 792 & 75.4 & 58.8 & 98.0 & 48.3 & 66.7 & 72.8 & 62.9 \\
\midrule
\multicolumn{10}{@{}l}{\emph{Number of offsets $K$, with $W{=}13$}} \\
8 & 13 & 104 & 10.0 & 58.7 & 93.4 & 44.3 & 71.9 & 76.1 & 60.3 \\
12 & 13 & 156 & 15.0 & 63.0 & 96.8 & 55.7 & 78.1 & 79.7 & 58.1 \\
16 & 13 & 208 & 20.0 & 65.3 & 97.4 & 57.0 & 80.2 & 80.6 & 64.1 \\
\textbf{24} & \textbf{13} & 312 & 29.9 & 68.8 & 99.1 & 62.0 & 85.7 & 86.0 & 68.5 \\
32 & 13 & 416 & 39.8 & 67.7 & 99.5 & 63.4 & 83.3 & 83.3 & 73.3 \\
48 & 13 & 624 & 59.7 & 71.3 & 99.5 & 68.7 & \best{86.2} & 88.1 & \best{73.4} \\
\midrule
\multicolumn{10}{@{}l}{\emph{Offsets instead of window length}} \\
32 & 5 & 160 & 15.4$^{*}$ & \best{75.8} & \best{99.9} & \best{78.3} & 82.2 & 87.1 & 70.9 \\
48 & 5 & 240 & 23.2 & 74.0 & 99.9 & 73.8 & 80.5 & 88.8 & 71.1 \\
48 & 9 & 432 & 41.3$^{*}$ & 71.1 & 99.7 & 73.3 & 83.3 & 86.9 & 73.4 \\
48 & 33 & 1584 & 150.4 & 64.3 & 98.2 & 52.1 & 74.3 & 78.0 & 68.4 \\
\bottomrule
\end{tabular}
\end{table*}

\paragraph{Profile normalization and visualization.}
Profiles use the frame-baseline residual in \cref{eq:profile}.
They exclude window position $j=0$ and are normalized by the mean $\bar E$ of the full reconstruction-error table.
For the phase-locking histograms, spectra and barcode views in \cref{fig:supp-phaselock,fig:supp-spectrum,fig:supp-barcode}, we also subtract the mean real-video profile at the same scan resolution.
This profile comes from 300 AIGVDBench reals for 480p and from the same batch of 60 DF26 reals for 720p.
The subtraction removes a shared offset-dependent bias of the measurement VAE, and the displayed profiles and the reported phase concentration $R$ use it.
Detection statistics and their AUROCs use the original normalized $r(k)$ without this subtraction.

For DF26 and the 2026 models, the 720p census rescans the same clips as the 480p census against the same 271 DF26 reals, so the two columns differ only in scan resolution.
The 720p Grok and Veo~3.1 groups of \cref{fig:supp-phaselock,fig:supp-spectrum,fig:supp-barcode} come from an earlier subset of 60 real clips, 40 Grok clips and 50 Veo clips, with the Veo clips restricted to a motion energy of at least 2.

\section{Training details}
\label{sec:supp-train}
\paragraph{Artifact path and \methodgf.}
Both use DINOv3 ViT-L/16 from the timm model \texttt{vit\_\allowbreak large\_\allowbreak patch16\_\allowbreak dinov3.\allowbreak lvd1689m}.
LoRA on the fused query--key--value projections has rank 16, $\alpha{=}32$ and dropout 0.05, which gives 1.6M trainable adapter parameters.
\methodgf uses one linear head with dropout 0.2 on the time-averaged final class token.
The artifact path uses the same generic head, trained with both members of each VAE pair labelled real, and an artifact head.
A measurement on 1{,}500 VAE pairs supports these counterfactual labels.
We measured detail as the log variance of the Laplacian and noise with Immerk{\ae}r's estimator~\cite{immerkaer}.
The reconstruction has less detail than its real counterpart in 98\% of pairs and less noise in 95\%.
Six other low-level statistics barely change.
The artifact head takes the class token and the patch-token mean and standard deviation of layers 8, 12 and 16, averaged over time.
These features pass through layer normalization, a 256-unit GELU layer, dropout 0.2 and a linear output, for 4.0M trainable parameters in total.
The two binary cross-entropy losses are summed with equal weights, and the gradient of the artifact head reaches the adapters.
Each head is standardized on the validation reals before the scores are summed.
We use AdamW with weight decay 0.05 and learning rates of $10^{-4}$ for the backbone adapters and $10^{-3}$ for the head.
Training runs for 8 epochs in bf16, with one warm-up epoch followed by cosine decay.
Each update uses 24 clips from batch size 8 and 3 accumulation steps.
Training augmentation uses RandomResizedCrop to 224 with scale 0.8--1 and horizontal flipping, and testing uses a center crop.
Like the baselines, the artifact path uses the final checkpoint.
\methodgf selects a checkpoint with held-out VAE pairs.
An artifact-path run takes about 1.5\,h on an H200.

\paragraph{Semantic path.}
We run Effort's released DeepfakeBench~\cite{deepfakebench} training code~\cite{effort} unchanged except for the data lists and the backbone.
The backbone is the same DINOv3 ViT-L/16 in fp32.
In each of its 24 blocks, we split the fused query--key--value projection into three $1024\times1024$ matrices.
Together with the output projections, this gives 96 matrices, and Effort's singular-value residual layers with $r{=}1023$ replace all of them.
The top 1023 singular components therefore stay frozen, and the smallest one is trained.
The replacement leaves the network unchanged at initialization, because on random inputs the class token changes by only a relative $3\times10^{-6}$.
All other weights are frozen.
The trainable parameters are the 96 residual components and a two-way linear head on the class token after the final normalization.
Together they have 0.20M parameters.
Frames are resized to $224\times224$ without cropping and normalized with ImageNet statistics.
Each of the 8 frames of a training clip is an independent sample, and batches of 32 frames receive Effort's eleven DeepfakeBench augmentations independently.
Training uses cross-entropy and Adam with learning rate $2\times10^{-4}$ and weight decay $5\times10^{-4}$, without a schedule or gradient clipping.
It runs in fp32 for 11 epochs, and we report the final epoch.
The score of a clip is the logit of its mean frame probability.
A run took 13--16\,h on a shared H200 and 24\,h on an A5000.

\paragraph{Fusion.}
We standardize the clip score of each path by its mean and standard deviation on the 3{,}000 AIGVDBench validation reals and sum the two standardized scores.
The same rule and calibration are used on every benchmark.
Both adapters can be merged into copies of the backbone weights, so inference costs two ViT-L forward passes.

\paragraph{Plug-in calibration.}
The plug-in calibrates $z(s_c)$ on 300 AIGVDBench validation reals that are scanned like the test clips.
It calibrates $z(s_d)$ on the detector's scores for all 3{,}000 AIGVDBench validation reals.
Both calibrations are reused on every benchmark.

\section{Post-processing chain}
\label{sec:supp-chain}
Each pair draws its VAE at random from the $4\times$ VAEs of Wan2.1, HunyuanVideo and CogVideoX and the $8\times$ VAE of LTX-Video.
Each VAE covers 24\% to 26\% of pairs.
The first 4{,}000 pairs use each source clip once and draw the reconstruction short side $G$ from $\{256,360,480,720\}$ with weights 4:2:2:1.
Another 1{,}333 pairs reuse source clips with $G=256$ for three quarters of them and $G=360$ for the rest.
They also use interpolation and duplication less often, and the probabilities below describe the first 4{,}000 pairs.
The 515 validation pairs come from 393 source clips that no training pair uses.
Both members of a pair share one parameter vector $\theta$.
It sets a leading trim of 0, 1, 2 or 3 frames and a uniformly sampled delivery short side $S\in\{480,720,1080\}$.
Motion-compensated interpolation is applied with probability 0.2 and a factor in $\{1.25,1.5,2\}$.
Duplicate-frame conversion is applied with probability 0.25 and excludes interpolation.
Decimation to 16\,fps is applied with probability 0.3 when applicable.
In duplication mode, both members first share a source decimated to 8--15\,fps.
Only the generated member undergoes the VAE round trip, and both members are returned to 24 or 30\,fps by repetition.
The encoding parameters are also shared.
The codec is x264 or x265 with equal probability, CRF is uniform in $[18,28]$, the number of B-frames is uniform in $[0,8]$, B-pyramid is enabled with probability 0.5, and the GOP length is uniform in $[12,250]$.
About 58\% of pairs undergo interpolation, duplication or decimation.
These operations can weaken pixel-domain periodicity, and the frame-level detector of \cref{sec:detector} learns the spatial decoding traces that remain.

\section{Baseline reproduction}
\label{sec:supp-baselines}
All baselines use the same 14{,}000 real and 14{,}000 Open-Sora training clips.
They also use the same 8 cached frames per clip at indices $0,4,\dots,28$ of a $256^2$ JPEG cache at quality 95.
The five 16-frame generators are AnimateDiff-T2V~\cite{animatediff}, SVD-I2V~\cite{svd}, SEINE-I2V~\cite{seine} and both VideoCrafter~\cite{videocrafter2} variants, and they use frames $0,2,\dots,14$.
An earlier version of the cache repeated their last frame, and all reported results use the corrected cache except the row marked $^*$ in \cref{tab:ablation}.
We keep method-specific augmentation, optimizers and schedules where they are compatible with the protocol.
We report the final checkpoint to avoid selection on in-domain validation scores, which are often saturated.
Official implementations are used throughout.

We adapted the protocol as follows.
Effort is trained with its released DeepfakeBench code, unchanged except for the data lists.
For DeMamba, fixed frames replace random temporal starts, and batch size 128 is implemented with $32\times4$ gradient accumulation.
Its X-CLIP~\cite{xclip} vision tower is loaded from the full pretrained X-CLIP model.
For ReStraV, the common 8-frame sequence replaces the official 24-frame trajectory at 12\,fps.
This gives 7 steps and 6 turning angles.
CNNSpot, Gram-Net and NPR are evaluated on full $256^2$ frames, following their official test commands.
The clip score of NPR is the logit of the mean frame probability computed in double precision, as \cref{sec:supp-numerics} explains.
UnivFD uses the AIGCDetectBenchmark settings with batch size 64 and augmentation probability 0.1.
AIGVDBench defines ForgeLens setting~1 as training on a small dataset for more epochs.
We report this setting and also the same schedule on the full training set.
Both ForgeLens settings use fp32 inference to avoid tied scores from autocast.
TimeSformer and UniFormerV2 use the official MMAction2 Kinetics-400 configurations with two output classes and a 50-epoch upper limit, so the cosine phase of UniFormerV2 is compressed to 45 epochs.
Native-Scale~\cite{nativescale} fine-tunes its Qwen2.5-VL vision encoder with the official training script and the learning rate of $10^{-5}$ from its paper.
The released step schedule would lower this rate to $10^{-6}$ from the first step.
These changes can affect performance, especially the change in temporal sampling, and they are part of the comparison protocol.

Reproduction gaps vary by method and generator split.
On closed-source generators, ForgeLens setting~3 spans 75.8\% to 87.1\% across seeds, against the reported 76.9\%.
Over three seeds, Effort reaches 84.3\% and 92.5\% on open-source and closed-source generators, compared with the reported 87.5\% and 94.1\%.
We keep the code revisions and per-video scores of each run.
For reference, AIGVDBench reports the following open-source and closed-source averages: CNNSpot 85.6\%/70.6\%, UnivFD 85.7\%/65.3\%, Gram-Net 80.6\%/65.7\%, FreDect 69.6\%/55.7\%, Fusing 79.5\%/65.5\%, NPR 68.8\%/52.4\%, DeMamba 81.0\%/69.4\%, DeCoF 82.6\%/72.9\%, TimeSformer 81.5\%/86.5\%, UniFormerV2 83.3\%/87.1\%, ForgeLens setting~3 89.6\%/76.9\%, ForgeLens setting~1 91.8\%/85.0\% and Effort 87.5\%/94.1\%.

\section{Additional results}
\label{sec:supp-results}
\paragraph{Test sets.}
\label{sec:supp-testsets}
AIGVDBench tests 3{,}000 clips per open-source generator and 2{,}000 per closed-source generator against 3{,}000 real clips.
DF26 has 271 real public-speaking clips and 2{,}420 generated clips from four closed-source models and Wan~2.2, HunyuanVideo~1.5 and LTX~2.3.
The 2026 models contribute 681 MiniMax~H3, 680 Seedance~2.5 and 953 Seedance~2.0 single-shot clips.
We compare them with the DF26 reals and average over the three models.
ViF-Bench has 165 real and 2{,}995 generated clips from 19 generators, and all of them are re-encoded to 256p at 16\,fps.

\paragraph{Generalization to newer generators.}
\Cref{tab:newgen} compares the zero-shot scan, \method and Effort with and without the plug-in, each over 3 seeds.
It also includes TimeSformer, which is the strongest other baseline on DF26's closed-source generators.
The zero-shot column scans at native resolution with a short side of at most 720 pixels.
It uses $\Cfour$ unless the family has another period, which means $\Ceight$ for Kling~3.0 and LTX~2.3 T2V and $\Ppk$ for Wan~2.6.
The 2026 models are single-shot clips from public collections~\cite{kivi,helioseedance}.
Only MiniMax~H3 discloses its VAE, which is a $4\times$ causal VAE~\cite{minimaxh3}, and Unknown denotes an unverified source VAE.
Over the four closed-source generators of DF26 and the three 2026 models, \method averages 83.0\%.
It exceeds TimeSformer by 2.6\%, and the plug-in adds another 2.5\%.
Among the 2026 models, \method is strongest on MiniMax~H3 and Seedance~2.5, and it exceeds Effort by 17\% on Seedance~2.5.
On Grok Imagine, the artifact path alone falls to 45.0\%, and the semantic path keeps \method above chance.
The plug-in adds 5\% to 7\% to \method on Grok, Kling~3.0 and Wan~2.6.
These results support transfer to newer generators.
\begin{table*}[t]
\centering
\caption{\textbf{Generalization to newer generators} (AUROC, \%) against the 271 DF26 reals. Bold marks the best per row, and the zero-shot column scans all clips at native $\le$720p as in the plug-in.}
\label{tab:newgen}
\small
\setlength{\tabcolsep}{5pt}
\begin{tabular}{@{}llc|cc|cc|c@{}}
\toprule
 & & zero-shot & \multicolumn{2}{c|}{\method} & \multicolumn{2}{c|}{Effort} & TimeS- \\
generator & VAE / period & scan & & + plug-in & & + plug-in & former \\
\midrule
\multicolumn{8}{@{}l}{\emph{DF26, closed-source}} \\
Veo 3.1 & Unknown / 4 & 73.4 & 77.9 & 78.2 & 77.1 & 77.8 & \best{83.6} \\
Kling 3.0 & Unknown / 8 & \best{88.2} & 73.2 & 77.9 & 57.9 & 66.5 & 52.8 \\
Grok Imagine & Unknown / 4 & 79.4 & 72.7 & 79.7 & 76.1 & \best{81.6} & 80.5 \\
Wan 2.6, 30 fps & Unknown / 5 & 96.0 & 92.5 & \best{97.9} & 82.6 & 96.9 & 86.4 \\
\multicolumn{8}{@{}l}{\emph{DF26, open-source}} \\
Wan 2.2 T2V & Wan $4\times$ / 4 & \best{98.8} & 83.3 & 98.1 & 76.2 & 97.6 & 80.1 \\
Wan 2.2 I2V & Wan $4\times$ / 4 & \best{96.7} & 85.1 & 96.0 & 68.3 & 90.0 & 57.4 \\
HunyuanVideo 1.5 T2V & own $4\times$ / 4 & \best{84.3} & 49.6 & 72.6 & 50.1 & 70.4 & 56.9 \\
HunyuanVideo 1.5 I2V & own $4\times$ / 4 & \best{81.9} & 77.4 & \best{81.9} & 60.9 & 70.2 & 48.0 \\
LTX 2.3 T2V & own $8\times$ / 8 & 74.3 & \best{82.0} & \best{82.0} & 75.0 & 76.5 & 66.0 \\
LTX 2.3 I2V & own $8\times$ / 8 & 60.0 & \best{91.7} & 87.9 & 79.7 & 76.8 & 58.8 \\
\multicolumn{8}{@{}l}{\emph{2026 closed-source models, against DF26 reals}} \\
MiniMax H3 & own $4\times$ / 4 & 77.7 & 90.0 & \best{90.5} & 88.0 & 89.1 & 88.0 \\
Seedance 2.5 & Unknown / 4 & 71.8 & \best{91.1} & 90.1 & 74.1 & 74.6 & 84.8 \\
Seedance 2.0 & Unknown / 4 & 78.0 & 83.7 & 84.2 & 82.8 & 83.7 & \best{86.4} \\
\bottomrule
\end{tabular}
\end{table*}

\paragraph{Extended ablations.}
\Cref{tab:ablation} extends \cref{tab:abl-rhythm,tab:abl-detector}.
Rows use seed 0 unless marked, and 3-seed rows average all seeds.
The row marked $^*$ was evaluated before the short-clip frame-cache fix described in \cref{sec:supp-baselines}.
Aggregate performance remains high when individual VAEs are excluded.
LTX is the only $8\times$ VAE, and removing it produces the largest drop on closed-source generators.
Within the artifact path, separating a generic head from the artifact head keeps the AIGVDBench performance over 3 seeds.
It also improves transfer by 6.2\% on DF26's closed-source generators and 4.3\% on the 2026 models.
Blocking the gradient of the artifact head to the backbone lowers the AIGVDBench performance.
Using Effort itself on CLIP-L/14 as the semantic path gives similar AIGVDBench results but 3.9\% less on DF26's closed-source generators, and it requires a second backbone.
We did not tune the path weights.
\begin{table}[tb]
\centering
\caption{\textbf{Extended ablations} (AUROC, \%) on AIGVDBench's open and closed averages and DF26's closed-source generators.}
\label{tab:ablation}
\small
\setlength{\tabcolsep}{3.5pt}
\begin{tabular}{@{}lcc@{}}
\toprule
 & AIGVD o/c & DF26 c \\
\midrule
\multicolumn{3}{@{}l}{\emph{(a) training data, DINOv3-L LoRA}} \\
benchmark set only & 68.9 / 81.2 & 65.2 \\
\ + post-processed reals, 3 seeds & 73.8 / 86.2 & 64.3 \\
\ + \pairs, single head, 3 seeds & 93.0 / 92.2 & 59.3 \\
real + pairs, \methodgf, 3 seeds & 90.9 / 91.9 & 51.3 \\
\midrule
\multicolumn{3}{@{}l}{\emph{(b) leave one VAE out, single head}} \\
$-$ Wan2.1 & 93.4 / 93.3 & 61.1 \\
$-$ CogVideoX & 91.9 / 91.9 & 54.4 \\
$-$ HunyuanVideo & 90.0 / 87.4 & 49.0 \\
$-$ LTX, $8\times$ & 89.1 / 82.8 & 35.7 \\
\midrule
\multicolumn{3}{@{}l}{\emph{(c) pairs on Effort}} \\
Effort, 3 seeds & 84.3 / 92.5 & 73.4 \\
\ + \pairs, 3 seeds & 93.1 / 77.3 & 47.6 \\
\midrule
\multicolumn{3}{@{}l}{\emph{(d) heads of the artifact path, 3 seeds}} \\
single head & 93.0 / 92.2 & 59.3 \\
generic + artifact head & 94.1 / 92.0 & 65.5 \\
\ no artifact-head grad.$^*$ & 89.5 / 84.7 & 38.6 \\
\midrule
\multicolumn{3}{@{}l}{\emph{(e) paths of \method, 3 seeds}} \\
artifact path only, as in (d) & 94.1 / 92.0 & 65.5 \\
semantic path only & 82.9 / 93.0 & 83.2 \\
both paths, \method & 93.9 / 95.8 & 79.0 \\
\ Effort on CLIP as semantic path & 94.4 / 96.3 & 75.1 \\
\bottomrule
\end{tabular}
\end{table}

\ifrobust
\paragraph{Post-processing robustness.}
\label{sec:supp-robust}
\Cref{fig:robustness} uses 100 real clips and 15 clips from each of the ten DF26 generators, sampled with a fixed seed.
We keep the first 2.5\,s of every clip, which covers the 31 frames that the detectors read and the 36 frames of a scan.
Re-compression re-encodes a clip with x264 at a CRF of 23, 28, 33 or 38 and keeps its size and frame rate.
Downscaling resizes the short side from 720 to 540 or 360 pixels with area interpolation and re-encodes at CRF 23.
Interpolation raises the frame rate by 1.25 or 2 times with the motion-compensated interpolation of FFmpeg and re-encodes at CRF 23.
Real and generated clips receive identical processing, while the calibration reals stay unprocessed.
The scans follow \cref{sec:plugin} and skip CRF 23 and 33 to save computation.
\method uses its seed-2 model, and TimeSformer uses seed 0.
At CRF 38 and a 360-pixel short side, the 95\% bootstrap intervals of the difference between \method and TimeSformer, both with the plug-in, include zero.
\fi

\paragraph{Swapping the real side.}
A restricted real-video source can confound generation artifacts with the appearance of that source, and audits report such confounds in existing benchmarks~\cite{fakejpeg,vidaudit}.
\Cref{tab:swapreal} tests this by swapping the real videos while the generated clips and trained detectors stay fixed.
An AUROC below 50\% means that genuine videos are ranked as more synthetic than the fakes.
With Youku clips~\cite{youkumplug} as reals, all trained detectors in the table degrade sharply.
The drop is therefore a shared problem of detectors trained on one real source rather than a weakness of a particular method.
In every swapped setting, \method stays above Effort and above the same backbone trained without pairs.
\methodgf is trained only on real videos and their pairs, and it is the most robust trained detector in all three settings.
The training-free D3~\cite{d3} improves instead.
Its values come from an earlier evaluation on our frame cache, so only its change, not its values, is comparable to its paper.
All detectors in \cref{tab:plugin-full} stay near chance on ViF-Bench, and a real-source shift explains this.
Against OpenVid reals, \method separates the generated clips of ViF-Bench with more than 96\% AUROC.
The difficulty comes from the ViF-Bench reals, which are re-encoded to 256p at 16\,fps.
Because the engine needs only real videos, it can add new real sources without generator clips.
The last row is a separately trained \methodgf that adds 2{,}166 disjoint Youku clips as a second real source and broadens the processing chain.
It is evaluated with 500 clips per generator against Youku reals, and a single-head detector trained in the same way with benchmark data reaches 70.4\% and 67.2\%.
This costs about 3\% on AIGVDBench under that protocol.
\begin{table*}[t]
\centering
\caption{\textbf{Swapping the real side} (open / closed AUROC, \%). Generated clips and models are fixed, and only the real videos change.}
\label{tab:swapreal}
\footnotesize
\setlength{\tabcolsep}{5pt}
\begin{tabular}{@{}lcccccccc@{}}
\toprule
real side & \method & artifact path & semantic path & \methodgf & no pairs & Effort & CNNSpot & D3 \\
\midrule
OpenVid, protocol & 93.9 / 95.8 & 94.1 / 92.0 & 82.9 / 93.0 & 90.9 / 91.9 & 68.9 / 81.2 & 84.3 / 92.5 & 82.1 / 71.4 & 45.6 / 62.7 \\
same clips, re-encoded 360p & 32.3 / 28.8 & 30.8 / 22.6 & 45.4 / 58.1 & 65.8 / 59.8 & 18.7 / 11.8 & 31.0 / 28.2 & 34.3 / 17.5 & 51.7 / 67.9 \\
Youku, GenVideo-100K & 17.4 / 11.9 & 17.9 / 10.7 & 24.8 / 30.8 & 45.2 / 34.5 & 12.2 / 5.4 & 14.6 / 8.5 & 19.9 / 8.5 & 78.4 / 86.6 \\
ViF-Bench reals & 28.9 / 25.4 & 27.9 / 21.0 & 35.2 / 42.6 & 54.9 / 46.4 & 20.8 / 16.8 & 21.7 / 17.7 & 28.0 / 15.7 & -- \\
\midrule
Youku, engine + Youku reals & -- & -- & -- & 71.2 / 57.5 & -- & -- & -- & -- \\
\bottomrule
\end{tabular}
\end{table*}

\paragraph{Plug-in, full table and sensitivity.}
\Cref{tab:plugin-full} reports detectors with and without the fixed fusion rule, including the two paths of \method and the training variants of \cref{tab:abl-detector}.
\Cref{tab:plugin-sens} compares the reported standardization with the original and pure rank standardizations.
\begin{table*}[t]
\centering
\caption{\textbf{The \plugin, full table} (AUROC, \%). Each detector appears with and without the plug-in, and the table includes extra training variants.}
\label{tab:plugin-full}
\scriptsize
\setlength{\tabcolsep}{3pt}
\begin{tabular}{@{}lcc|c|c|cccc@{}}
\toprule
 & \multicolumn{2}{c|}{AIGVDBench} & 2026 closed & ViF & \multicolumn{4}{c@{}}{DF26} \\
Detector & open & closed & & & all & open & closed & 2AFC \\
\midrule
zero-shot rhythm alone ($\Ppk$) & 69.0 & 63.0 & 62.5 & 56.9 & 76.2 & 74.4 & 79.0 & 76.1 \\
\midrule
CNNSpot & 82.9 & 71.6 & 78.0 & 47.0 & 53.5 & 65.5 & 35.5 & 61.0 \\
\quad + \plugin & 86.4 \dsc{+3.5} & 76.9 \dsc{+5.3} & 80.3 \dsc{+2.3} & 51.4 \dsc{+4.4} & 69.8 \dsc{+16.3} & 79.5 \dsc{+14.0} & 55.3 \dsc{+19.8} & 75.1 \dsc{+14.2} \\
Gram-Net & 83.3 & 67.8 & 76.9 & 52.4 & 52.9 & 63.3 & 37.3 & 59.3 \\
\quad + \plugin & 86.1 \dsc{+2.8} & 74.7 \dsc{+6.9} & 79.2 \dsc{+2.3} & 56.8 \dsc{+4.4} & 71.0 \dsc{+18.1} & 79.2 \dsc{+15.9} & 58.8 \dsc{+21.5} & 74.5 \dsc{+15.2} \\
FreDect & 59.3 & 53.6 & 60.8 & 49.9 & 60.8 & 59.0 & 63.4 & 62.4 \\
\quad + \plugin & 71.9 \dsc{+12.7} & 65.2 \dsc{+11.7} & 66.4 \dsc{+5.6} & 54.8 \dsc{+4.9} & 74.7 \dsc{+13.9} & 74.3 \dsc{+15.3} & 75.3 \dsc{+11.9} & 76.9 \dsc{+14.5} \\
Fusing & 64.6 & 51.3 & 64.8 & 45.4 & 46.0 & 53.3 & 35.0 & 46.4 \\
\quad + \plugin & 74.7 \dsc{+10.1} & 63.7 \dsc{+12.4} & 70.2 \dsc{+5.4} & 50.4 \dsc{+5.1} & 68.3 \dsc{+22.3} & 74.2 \dsc{+20.9} & 59.5 \dsc{+24.5} & 69.6 \dsc{+23.2} \\
UnivFD & 80.7 & 64.8 & 60.3 & 47.5 & 54.0 & 59.4 & 46.1 & 62.5 \\
\quad + \plugin & 85.0 \dsc{+4.3} & 71.8 \dsc{+7.1} & 65.5 \dsc{+5.2} & 54.1 \dsc{+6.5} & 69.1 \dsc{+15.1} & 74.4 \dsc{+15.0} & 61.2 \dsc{+15.2} & 73.9 \dsc{+11.4} \\
NPR & 73.9 & 61.4 & 69.5 & 45.4 & 44.1 & 53.1 & 30.7 & 47.2 \\
\quad + \plugin & 80.9 \dsc{+7.0} & 70.9 \dsc{+9.4} & 73.6 \dsc{+4.2} & 50.8 \dsc{+5.5} & 67.9 \dsc{+23.7} & 74.1 \dsc{+21.1} & 58.5 \dsc{+27.7} & 70.6 \dsc{+23.5} \\
TimeSformer & 74.8 & 91.2 & 86.4 & 53.7 & 67.0 & 61.2 & 75.8 & 73.4 \\
\quad + \plugin & 81.2 \dsc{+6.4} & 90.1 \dsc{$-$1.1} & 85.5 \dsc{$-$0.9} & 56.3 \dsc{+2.6} & 77.8 \dsc{+10.7} & 74.9 \dsc{+13.7} & 82.1 \dsc{+6.3} & 80.9 \dsc{+7.5} \\
UniFormerV2 & 81.1 & 92.1 & 64.9 & 48.9 & 50.6 & 47.5 & 55.3 & 52.4 \\
\quad + \plugin & 85.5 \dsc{+4.3} & 92.0 \dsc{$-$0.2} & 69.5 \dsc{+4.6} & 50.6 \dsc{+1.7} & 68.6 \dsc{+18.0} & 67.4 \dsc{+20.0} & 70.4 \dsc{+15.1} & 70.9 \dsc{+18.5} \\
DeMamba & 81.2 & 78.5 & 71.6 & 48.0 & 61.4 & 60.8 & 62.3 & 67.2 \\
\quad + \plugin & 85.7 \dsc{+4.5} & 82.1 \dsc{+3.6} & 75.2 \dsc{+3.6} & 51.0 \dsc{+2.9} & 74.5 \dsc{+13.1} & 74.9 \dsc{+14.1} & 74.0 \dsc{+11.6} & 78.2 \dsc{+10.9} \\
DeCoF & 84.8 & 78.6 & 70.9 & 53.5 & 59.8 & 60.7 & 58.5 & 67.9 \\
\quad + \plugin & 87.8 \dsc{+3.0} & 80.8 \dsc{+2.2} & 73.6 \dsc{+2.7} & 58.1 \dsc{+4.6} & 73.2 \dsc{+13.4} & 75.5 \dsc{+14.8} & 69.7 \dsc{+11.2} & 78.4 \dsc{+10.5} \\
ReStraV & 49.9 & 46.5 & 62.7 & 57.6 & 57.4 & 59.7 & 53.9 & 55.5 \\
\quad + \plugin & 63.3 \dsc{+13.4} & 58.7 \dsc{+12.2} & 66.1 \dsc{+3.4} & 60.6 \dsc{+3.0} & 70.7 \dsc{+13.3} & 72.2 \dsc{+12.5} & 68.4 \dsc{+14.5} & 69.8 \dsc{+14.2} \\
ForgeLens, setting 3 & 74.5 & 83.1 & 74.3 & 40.2 & 55.6 & 52.7 & 60.1 & 60.0 \\
\quad + \plugin & 81.8 \dsc{+7.2} & 85.2 \dsc{+2.1} & 78.1 \dsc{+3.8} & 44.0 \dsc{+3.8} & 74.2 \dsc{+18.5} & 73.8 \dsc{+21.1} & 74.8 \dsc{+14.7} & 76.6 \dsc{+16.6} \\
ForgeLens, setting 1 & 76.0 & 65.6 & 63.6 & 38.6 & 46.6 & 56.9 & 31.3 & 50.2 \\
\quad + \plugin & 82.7 \dsc{+6.7} & 72.9 \dsc{+7.3} & 68.4 \dsc{+4.8} & 42.0 \dsc{+3.3} & 66.0 \dsc{+19.3} & 75.0 \dsc{+18.1} & 52.4 \dsc{+21.1} & 68.5 \dsc{+18.3} \\
\quad setting-1 schedule, full data & 86.1 & 88.2 & 83.2 & 44.4 & 60.3 & 62.8 & 56.5 & 66.9 \\
\quad + \plugin & 88.8 \dsc{+2.7} & 87.8 \dsc{$-$0.4} & 85.2 \dsc{+2.0} & 46.5 \dsc{+2.1} & 75.4 \dsc{+15.1} & 78.9 \dsc{+16.1} & 70.1 \dsc{+13.6} & 79.9 \dsc{+13.1} \\
Effort & 85.6 & 92.9 & 81.6 & 50.5 & 70.4 & 68.4 & 73.4 & 82.6 \\
\quad + \plugin & 87.9 \dsc{+2.3} & 92.2 \dsc{$-$0.8} & 82.5 \dsc{+0.8} & 53.7 \dsc{+3.3} & 80.4 \dsc{+10.0} & 80.2 \dsc{+11.9} & 80.7 \dsc{+7.3} & 87.1 \dsc{+4.4} \\
Native-Scale & 74.1 & 68.3 & 78.9 & 47.8 & 52.2 & 53.9 & 49.6 & 57.9 \\
\quad + \plugin & 80.8 \dsc{+6.7} & 75.1 \dsc{+6.8} & 81.3 \dsc{+2.4} & 50.7 \dsc{+2.8} & 71.4 \dsc{+19.2} & 73.5 \dsc{+19.5} & 68.3 \dsc{+18.8} & 75.2 \dsc{+17.3} \\
DINOv3-L LoRA, no pairs & 66.8 & 81.5 & 79.3 & 43.2 & 62.1 & 60.0 & 65.2 & 70.3 \\
\quad + \plugin & 76.4 \dsc{+9.5} & 84.2 \dsc{+2.7} & 82.5 \dsc{+3.2} & 44.2 \dsc{+0.9} & 76.5 \dsc{+14.4} & 76.4 \dsc{+16.4} & 76.5 \dsc{+11.4} & 82.0 \dsc{+11.6} \\
\methodgf & 92.0 & 91.9 & 84.8 & 49.1 & 65.8 & 75.5 & 51.3 & 70.5 \\
\quad + \plugin & 92.7 \dsc{+0.7} & 91.6 \dsc{$-$0.3} & 85.2 \dsc{+0.4} & 52.4 \dsc{+3.3} & 75.4 \dsc{+9.6} & 85.1 \dsc{+9.6} & 60.9 \dsc{+9.6} & 78.7 \dsc{+8.2} \\
\method & 95.0 & 96.4 & 88.2 & 47.0 & 78.5 & 78.2 & 79.0 & 87.7 \\
\quad + \plugin & 93.8 \dsc{$-$1.2} & 94.6 \dsc{$-$1.8} & 88.3 \dsc{+0.1} & 48.2 \dsc{+1.2} & 85.2 \dsc{+6.7} & 86.4 \dsc{+8.2} & 83.5 \dsc{+4.4} & 89.2 \dsc{+1.4} \\
\method, artifact path only & 95.4 & 92.2 & 89.1 & 45.4 & 73.5 & 78.8 & 65.5 & 80.4 \\
\quad + \plugin & 94.2 \dsc{$-$1.2} & 91.4 \dsc{$-$0.8} & 89.2 \dsc{+0.1} & 46.7 \dsc{+1.3} & 81.5 \dsc{+8.0} & 86.6 \dsc{+7.8} & 73.8 \dsc{+8.3} & 85.3 \dsc{+5.0} \\
\method, semantic path only & 83.2 & 93.5 & 76.5 & 55.5 & 74.3 & 68.3 & 83.2 & 86.1 \\
\quad + \plugin & 86.2 \dsc{+3.0} & 92.9 \dsc{$-$0.6} & 78.8 \dsc{+2.4} & 59.2 \dsc{+3.6} & 83.5 \dsc{+9.3} & 81.1 \dsc{+12.8} & 87.2 \dsc{+4.0} & 88.5 \dsc{+2.5} \\
artifact path, single head & 94.5 & 92.2 & 84.8 & 47.6 & 70.4 & 77.7 & 59.3 & 75.1 \\
\quad + \plugin & 94.5 \dsc{+0.0} & 92.6 \dsc{+0.5} & 85.0 \dsc{+0.2} & 49.7 \dsc{+2.1} & 78.6 \dsc{+8.2} & 86.0 \dsc{+8.2} & 67.6 \dsc{+8.3} & 81.1 \dsc{+6.0} \\
DINOv3-L LoRA, post-processed reals only & 72.0 & 87.1 & 82.6 & 45.3 & 60.7 & 58.3 & 64.3 & 68.0 \\
\quad + \plugin & 80.3 \dsc{+8.3} & 89.3 \dsc{+2.1} & 85.2 \dsc{+2.7} & 46.5 \dsc{+1.2} & 75.0 \dsc{+14.3} & 75.3 \dsc{+17.0} & 74.6 \dsc{+10.3} & 79.4 \dsc{+11.5} \\
Effort + pairs & 94.6 & 78.4 & 85.1 & 49.2 & 68.4 & 82.3 & 47.5 & 73.5 \\
\quad + \plugin & 94.5 \dsc{$-$0.1} & 80.6 \dsc{+2.2} & 85.4 \dsc{+0.3} & 52.9 \dsc{+3.6} & 77.0 \dsc{+8.6} & 89.1 \dsc{+6.8} & 58.9 \dsc{+11.3} & 79.5 \dsc{+6.1} \\
\bottomrule
\end{tabular}
\end{table*}

\begin{table*}[t]
\centering
\caption{\textbf{Plug-in sensitivity} (AUROC, \%). The reported rule is compared with the original and pure-rank standardizations on the three columns where fusion can hurt a host.}
\label{tab:plugin-sens}
\scriptsize
\setlength{\tabcolsep}{2.6pt}
\begin{tabular}{@{}lcccc|cccc|cccc@{}}
\toprule
 & \multicolumn{4}{c|}{AIGVDBench closed} & \multicolumn{4}{c|}{2026 closed} & \multicolumn{4}{c@{}}{DF26 closed} \\
Detector & none & reported & original & rank & none & reported & original & rank & none & reported & original & rank \\
\midrule
\multicolumn{13}{@{}l}{\emph{Image forensics, frame-level}} \\
CNNSpot & 71.6 & 76.9 \dsc{+5.3} & 77.0 \dsc{+5.3} & 76.9 \dsc{+5.3} & 78.0 & 80.3 \dsc{+2.3} & 80.2 \dsc{+2.2} & 80.3 \dsc{+2.3} & 35.5 & 55.3 \dsc{+19.8} & 55.2 \dsc{+19.7} & 55.3 \dsc{+19.8} \\
Gram-Net & 67.8 & 74.7 \dsc{+6.9} & 73.9 \dsc{+6.1} & 74.7 \dsc{+6.9} & 76.9 & 79.2 \dsc{+2.3} & 77.2 \dsc{+0.3} & 79.2 \dsc{+2.3} & 37.3 & 58.8 \dsc{+21.5} & 60.9 \dsc{+23.6} & 58.8 \dsc{+21.5} \\
FreDect & 53.6 & 65.2 \dsc{+11.7} & 64.3 \dsc{+10.8} & 65.2 \dsc{+11.7} & 60.8 & 66.4 \dsc{+5.6} & 67.2 \dsc{+6.5} & 66.4 \dsc{+5.6} & 63.4 & 75.3 \dsc{+11.9} & 78.8 \dsc{+15.4} & 75.3 \dsc{+11.9} \\
Fusing & 51.3 & 63.7 \dsc{+12.4} & 65.5 \dsc{+14.2} & 63.7 \dsc{+12.4} & 64.8 & 70.2 \dsc{+5.4} & 71.5 \dsc{+6.6} & 70.2 \dsc{+5.4} & 35.0 & 59.5 \dsc{+24.5} & 69.7 \dsc{+34.7} & 59.5 \dsc{+24.5} \\
UnivFD & 64.8 & 71.8 \dsc{+7.1} & 71.5 \dsc{+6.7} & 71.8 \dsc{+7.1} & 60.3 & 65.5 \dsc{+5.2} & 64.1 \dsc{+3.9} & 65.5 \dsc{+5.3} & 46.1 & 61.2 \dsc{+15.2} & 61.8 \dsc{+15.8} & 61.2 \dsc{+15.2} \\
NPR & 61.4 & 70.9 \dsc{+9.4} & 70.2 \dsc{+8.7} & 70.9 \dsc{+9.4} & 69.5 & 73.6 \dsc{+4.2} & 72.1 \dsc{+2.7} & 73.7 \dsc{+4.2} & 30.7 & 58.5 \dsc{+27.7} & 60.9 \dsc{+30.1} & 58.5 \dsc{+27.7} \\
\multicolumn{13}{@{}l}{\emph{Video classifiers}} \\
TimeSformer & 91.2 & 90.1 \dsc{$-$1.1} & 82.5 \dsc{$-$8.8} & 90.0 \dsc{$-$1.2} & 86.4 & 85.5 \dsc{$-$0.9} & 76.6 \dsc{$-$9.8} & 85.5 \dsc{$-$0.9} & 75.8 & 82.1 \dsc{+6.3} & 80.2 \dsc{+4.3} & 82.1 \dsc{+6.3} \\
UniFormerV2 & 92.1 & 92.0 \dsc{$-$0.2} & 79.4 \dsc{$-$12.7} & 92.0 \dsc{$-$0.2} & 64.9 & 69.5 \dsc{+4.6} & 67.3 \dsc{+2.5} & 69.5 \dsc{+4.6} & 55.3 & 70.4 \dsc{+15.1} & 76.1 \dsc{+20.8} & 70.4 \dsc{+15.1} \\
\multicolumn{13}{@{}l}{\emph{Video detectors}} \\
DeMamba & 78.5 & 82.1 \dsc{+3.6} & 73.3 \dsc{$-$5.2} & 82.1 \dsc{+3.6} & 71.6 & 75.2 \dsc{+3.6} & 71.2 \dsc{$-$0.4} & 75.3 \dsc{+3.7} & 62.3 & 74.0 \dsc{+11.6} & 76.2 \dsc{+13.8} & 74.1 \dsc{+11.7} \\
DeCoF & 78.6 & 80.8 \dsc{+2.2} & 78.2 \dsc{$-$0.4} & 80.8 \dsc{+2.2} & 70.9 & 73.6 \dsc{+2.7} & 71.3 \dsc{+0.4} & 73.6 \dsc{+2.7} & 58.5 & 69.7 \dsc{+11.2} & 71.1 \dsc{+12.6} & 69.7 \dsc{+11.2} \\
ReStraV & 46.5 & 58.7 \dsc{+12.2} & 58.7 \dsc{+12.2} & 58.7 \dsc{+12.2} & 62.7 & 66.1 \dsc{+3.4} & 64.8 \dsc{+2.1} & 66.1 \dsc{+3.4} & 53.9 & 68.4 \dsc{+14.5} & 70.9 \dsc{+17.0} & 68.4 \dsc{+14.5} \\
ForgeLens, setting 3 & 83.1 & 85.2 \dsc{+2.1} & 73.2 \dsc{$-$9.9} & 85.2 \dsc{+2.0} & 74.3 & 78.1 \dsc{+3.8} & 70.2 \dsc{$-$4.2} & 78.1 \dsc{+3.8} & 60.1 & 74.8 \dsc{+14.7} & 76.8 \dsc{+16.8} & 74.8 \dsc{+14.7} \\
ForgeLens, setting 1 & 65.6 & 72.9 \dsc{+7.3} & 68.7 \dsc{+3.1} & 72.9 \dsc{+7.3} & 63.6 & 68.4 \dsc{+4.8} & 67.0 \dsc{+3.4} & 68.4 \dsc{+4.8} & 31.3 & 52.4 \dsc{+21.1} & 69.5 \dsc{+38.2} & 52.4 \dsc{+21.2} \\
\quad setting-1 schedule, full data & 88.2 & 87.8 \dsc{$-$0.4} & 74.4 \dsc{$-$13.8} & 87.8 \dsc{$-$0.4} & 83.2 & 85.2 \dsc{+2.0} & 72.9 \dsc{$-$10.2} & 85.1 \dsc{+2.0} & 56.5 & 70.1 \dsc{+13.6} & 76.0 \dsc{+19.5} & 70.1 \dsc{+13.6} \\
Effort & 92.9 & 92.2 \dsc{$-$0.8} & 92.7 \dsc{$-$0.2} & 92.2 \dsc{$-$0.8} & 81.6 & 82.5 \dsc{+0.8} & 82.5 \dsc{+0.9} & 82.5 \dsc{+0.8} & 73.4 & 80.7 \dsc{+7.3} & 80.2 \dsc{+6.8} & 80.7 \dsc{+7.3} \\
Native-Scale & 68.3 & 75.1 \dsc{+6.8} & 75.3 \dsc{+7.0} & 75.1 \dsc{+6.8} & 78.9 & 81.3 \dsc{+2.4} & 81.6 \dsc{+2.7} & 81.3 \dsc{+2.4} & 49.6 & 68.3 \dsc{+18.8} & 67.5 \dsc{+18.0} & 68.3 \dsc{+18.8} \\
\multicolumn{13}{@{}l}{\emph{Ours}} \\
DINOv3-L LoRA, no pairs & 81.5 & 84.2 \dsc{+2.7} & 83.7 \dsc{+2.2} & 84.2 \dsc{+2.6} & 79.3 & 82.5 \dsc{+3.2} & 80.8 \dsc{+1.4} & 82.5 \dsc{+3.1} & 65.2 & 76.5 \dsc{+11.4} & 75.3 \dsc{+10.1} & 76.5 \dsc{+11.4} \\
\methodgf & 91.9 & 91.6 \dsc{$-$0.3} & 92.5 \dsc{+0.6} & 91.5 \dsc{$-$0.4} & 84.8 & 85.2 \dsc{+0.4} & 85.6 \dsc{+0.8} & 85.2 \dsc{+0.4} & 51.3 & 60.9 \dsc{+9.6} & 60.4 \dsc{+9.1} & 61.0 \dsc{+9.7} \\
\method & 96.4 & 94.6 \dsc{$-$1.8} & 95.7 \dsc{$-$0.7} & 94.5 \dsc{$-$1.8} & 88.2 & 88.3 \dsc{+0.1} & 88.9 \dsc{+0.6} & 88.4 \dsc{+0.2} & 79.0 & 83.5 \dsc{+4.4} & 82.4 \dsc{+3.4} & 83.7 \dsc{+4.6} \\
\bottomrule
\end{tabular}
\end{table*}

\paragraph{Plug-in variants.}
The reported rule uses $\tau{=}0$ and the quantile-based standardization of \cref{sec:plugin}.
We chose $\Ppk$ after inspecting the DF26 scans, including the period-5 signature of Wan~2.6, and then fixed it for the cross-detector evaluation.
The originally specified rule standardized detector scores by the mean and standard deviation of their real-video scores.
It is unstable for detectors with saturated probability outputs, because over 99\% of their validation-real scores are below $10^{-3}$ and their standard deviation is below 0.005.
Moderately confident fakes are then standardized to values comparable to the rhythm term of real videos.
For TimeSformer, UniFormerV2, DeMamba, ForgeLens setting~3 and the setting-1 schedule on the full data, it lowers the closed-source AUROC of AIGVDBench by up to 13.8\%, as \cref{tab:plugin-sens} lists, so we replaced it with the reported standardization.
The new rule approximately agrees with the old one for well-behaved scores inside the real range, and it keeps their scale beyond that range.
The pure rank standardization in \cref{tab:plugin-sens} maps $s_d$ to $\Phi^{-1}$ of its mid-rank quantile among the real videos, so it discards the detector's ordering of the clips above the calibration range.

\section{Numerical stability of score calibration}
\label{sec:supp-numerics}
Averaging sigmoid frame probabilities in single precision makes many NPR scores underflow to exactly zero, which ties real and generated clips.
All reported NPR results, including the plug-in, therefore use the logit of the mean frame probability computed in double precision.
This preserves the ordering of the probability mean without underflow.
For all detectors, paired 2AFC counts ties as one half, and detector heads are evaluated in fp32 to avoid quantized scores.
\makeatletter\setlength{\@fptop}{0pt}\setlength{\@dblfptop}{0pt}\makeatother

\end{document}